\documentclass[final]{cvpr}

\usepackage{times}
\usepackage{epsfig}
\usepackage{graphicx}
\usepackage{amsmath}
\usepackage{amssymb}
\usepackage{multirow}
\usepackage{pifont}
\usepackage{booktabs}
\usepackage{epsfig}
\usepackage{booktabs,multirow}
\usepackage{graphics}
\usepackage{threeparttable}
\usepackage{color}
\usepackage[normalem]{ulem}
\usepackage{multirow}
\usepackage{float}
\usepackage{amsfonts}
\usepackage{bm}
\usepackage{subfig}
\usepackage{enumitem}
\usepackage[numbers]{natbib}
\usepackage{mathtools}
\usepackage{marvosym}
\usepackage{array}
\usepackage[table]{xcolor}
\usepackage{colortbl}
\definecolor{bblue}{rgb}{0,150,230}
\definecolor{mygray}{gray}{.9}
\definecolor{myy}{RGB}{126,95,0}
\definecolor{ggray}{rgb}{160,160,160}

\newcolumntype{I}{!{\vrule width 1pt}}

\definecolor{ggray}{RGB}{127,127,127}

\makeatletter
\newcommand{\thickhline}{%
    \noalign {\ifnum 0=`}\fi \hrule height 1pt
    \futurelet \reserved@a \@xhline
}
\makeatother
\newcommand{\tabincell}[2]{\begin{tabular}{@{}#1@{}}#2\end{tabular}}

\usepackage{caption}
\usepackage[pagebackref=true,breaklinks=true,letterpaper=true,colorlinks,bookmarks=false]{hyperref}

\usepackage[utf8]{inputenc}

\usepackage{cleveref}
\crefname{section}{§}{§§}
\Crefname{section}{§}{§§}

\def\cvprPaperID{6136} 
\def\confYear{CVPR 2021}

\def\ourdataset{$\textit{FFIW}_{10K\!}$}
\begin{document}

\title{Face Forensics in the Wild}

\author{Tianfei Zhou$^{1}$~, Wenguan Wang$^{1}$\thanks{Corresponding author: \textit{Wenguan Wang}.}~~, Zhiyuan Liang$^2$~, Jianbing Shen$^{3,2}$\\
\small{$^1$ETH Zurich~~~$^2$Beijing Institute of Technology~~~$^3$Inception Institute of Artificial Intelligence}\\
\small\url{https://github.com/tfzhou/FFIW}
}

%

\maketitle
\thispagestyle{empty}


\begin{abstract}
On existing public benchmarks, face forgery detection techniques have achieved great success. However, when used in multi-person videos, which often contain many people active in the scene with only a small subset having been manipulated, their performance remains far from being satisfactory. To take face forgery detection to a new level, we construct a novel large-scale dataset, called \ourdataset, which comprises 10,000 high-quality forgery videos, with an average of three human faces in each frame. The manipulation procedure is fully automatic, controlled by a domain-adversarial quality assessment network, making our dataset highly scalable with low human cost. In addition, we propose a novel algorithm to tackle the task of multi-person face forgery detection. Supervised by only video-level label, the algorithm explores  multiple instance learning and learns to automatically attend to tampered faces. Our algorithm outperforms representative approaches for both forgery classification and localization on \ourdataset~\!, and also shows high generalization ability on existing benchmarks. We hope that our dataset and study will help the community to explore this new field in more depth.

\end{abstract}


\section{Introduction}
The rise of synthetic audiovisual media is forcing us towards a critical and unsettling realization: our belief that video and audio recordings are reliable representations of reality is no longer tenable. In particular, since emerging in 2017, the deepfake phenomenon has grown rapidly, requiring \textit{face forensics} to recognize potentially manipulated facial regions in images and videos. Accurate face forgery detection\footnote{In this work, ``forgery'' refers to altering imagery by swapping faces.} would have an immediate and
far-reaching impact in alleviating the malicious intents of deepfakes, such as, face
recognition attacks~\cite{korshunov2018deepfakes} and fake news~\cite{huh2018fighting,agarwal2019protecting}.

To help with this, several benchmarks have been established. The pioneering large-scale dataset, \ie, FaceForensics++\!~\cite{rossler2019faceforensics}, has greatly contributed to spurring interest and progress in the area of face forgery detection. However, as algorithms evolve, there have been signs of performance saturation on this dataset\!~\cite{qian2020thinking}. More recent datasets (\eg, DeeperForensics-1.0~\cite{jiang2020deeperforensics10}, DFDC~\cite{dolhansky2019deepfake}, Celeb-DF\!~\cite{li2020celeb}) thus employ more advanced synthesis techniques to produce highly realistic tampered faces. Even so, all previous datasets are subject to a significant limitation: they have a strong selection bias\!~\cite{torralba2011unbiased} to favor trimmed videos, each of which involves only one person. Thus, they provide insufficient representation of true visual world, which makes them inappropriate and unreliable to evaluate face forgery detection models in real-world, multi-person circumstances.

\begin{figure}[t]
	\begin{center}
		\includegraphics[width=\linewidth]{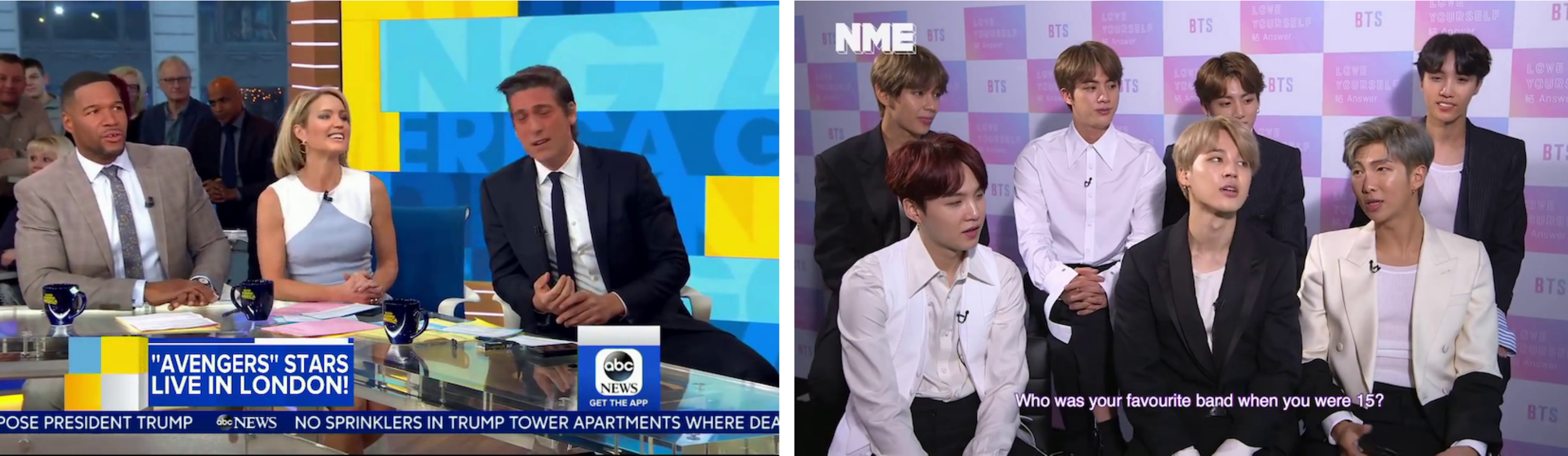}
		\vspace{-19pt}
		\captionsetup{font=small}
	\caption{\small\textbf{Representative examples from \ourdataset~\!, showing multi-person video frames with only a few faces being forged.} Can you recognize the manipulated ones?\protect\footnotemark}
		\label{fig:1}
	\end{center}%
	\vspace{-16pt}
\end{figure}

\footnotetext{Answer: The middle person in the left image and the rightmost of the back row in the right image are fake.}




To take the research of face forensics into a new level, we introduce a new large-scale dataset, called \ourdataset,  promoting empirical study of face forgery detection in multi-person scenarios. In \ourdataset, \textit{each video involves multiple individuals but only some, not all, faces are manipulated} (see \figref{fig:1}). This raises a significant challenge to current techniques: even for the fake videos, real faces are still in the majority.  In particular, \ourdataset~features $10,\!000$ high-fidelity manipulated videos, with $12$ seconds long on average, resulting in $33$ hours of video in total. In comparison with existing datasets, \ourdataset~has several distinguished features:$_{\!}$ \textbf{i)}$_{\!}$ \textit{Real-world}$_{\!}$ \textit{complexity.}$_{\!}$ The$_{\!}$ number of identities in each frame ranges from one to fifteen, with three on average, yielding a better representation of real visual scenes. \textbf{ii)} \textit{High} \textit{fidelity with low human cost.} The synthesis quality is controlled by a quality assessment network (Q-Net), which provides an effortless way to measure the realism of forged faces. \textbf{iii)} \textit{Large} \textit{scale.} \ourdataset~is comparable to the largest current dataset~\cite{jiang2020deeperforensics10} in terms of the number of unique fake videos (see Table~\ref{table:dataset}), and, more importantly, provides videos under multi-person settings.

\ourdataset~provides both video- and face-level annotations, allowing benchmarking methods on both forgery classification and localization tasks. In addition, to bring the research into a more natural setting, \ourdataset~provides two benchmark settings. In the first setting, benchmarking methods can make use of face-level supervision. However, in the second one, only video-level labels are allowed to be accessed during training. This makes the task more valuable from both academic and practical perspectives.

Along with \ourdataset, we propose a discriminative attention model for face forgery classification and localization in multi-person scenarios. The model explores the idea of multiple instance learning~\cite{dietterich1997solving,maron1998framework} and can be trained with video-level label only. It comprises three essential parts: i) a \textit{multi-temporal-scale instance feature aggregation} module that summarizes short-term, long-term and global features of each face tracklet to obtain a robust and discriminative representation; ii) an \textit{attention-based bag feature aggregation} module that adaptively aggregates the representations of all face tracklets into a video-level representation; and iii) a \textit{sparse regularization loss} to enforce the sparsity of face selection for real/fake discrimination. The sparsity-regularized attention learning mechanism is tasked with automatically selecting possible tampered faces during classification, promoting the localization ability of the system.




In summary, our contributions are three-fold:
\textbf{i)} To pave the avenue for face forgery detection in open world, we contribute \ourdataset~dataset to the community, which is distinctive in its real-world complexity. As far we know, it is the first large-scale face forensics dataset for \textit{fully unconstrained}, \textit{multi-person} face forgery detection.
\textbf{ii)} We propose a model-agnostic quality assessment model for synthesis quality management. Trained independently of deepfake methods, the model has high flexibility and accessibility, and can facilitate future dataset construction.
\textbf{iii)} We propose a discriminative attention model for multi-person face forgery detection. By revisiting multiple instance learning and gathering diverse temporal context, it provides promising performance on both fake video classification and fake face localization tasks with only video-level supervision.  

\begin{table}[t]
	\centering
	\small
	\resizebox{0.49\textwidth}{!}{
		\setlength\tabcolsep{1pt}
		\renewcommand\arraystretch{1.05}
		\begin{tabular}{r||cccccc}
			\hline\thickhline
			\rowcolor{mygray}
			&  &  &  & & \#Synthetic  & \#Face\\
			\rowcolor{mygray}
			\multirow{-2}{*}{Dataset} & \multirow{-2}{*}{Year} & \multirow{-2}{*}{Pub.} & \multirow{-2}{*}{\#Real} & \multirow{-2}{*}{\#Fake} & Methods   &Per-frame  \\ \hline\hline
			UADFV~\cite{yang2019exposing} & 2018 & ICASSP  &49 & 49 & 1 &  1  \\
			DeepFake-TIMIT~\cite{korshunov2018deepfakes} & 2018 & arXiv & 320 & 640 & 2 &   1 \\
			Deep Fake Detection~\cite{Google2019DeepFakeDetection} & 2019 &- & 363 & 3,068 & 5 &  1 \\
			FaceForenscics++~\cite{rossler2019faceforensics} & 2019 & ICCV& 1,000 & 4,000 & 4  & 1 \\
			DFDC Preview~\cite{dolhansky2019deepfake} & 2019 & arXiv & 1,131  & 4,113 & 2  &  $\sim$1  \\
			Celeb-DF~\cite{li2020celeb}  & 2020 & CVPR & 590 & 5,639 & 1 &  1 \\
			DeeperForensics-1.0~\cite{jiang2020deeperforensics10} & 2020 & CVPR & 50,000 & 10,000$^\dagger$ & 1  & 1 \\ \hline
			\textbf{\ourdataset} (Ours) & 2020 & - & 10,000 & 10,000 & 3 &  3.15 \\ \hline
		\end{tabular}
	}
	\captionsetup{font=small}
	\caption{\small\textbf{Comparisons of \ourdataset~with existing datasets.} As far as we know, \ourdataset~is the first specializing in \textit{unconstrained}, \textit{multi-person} face forgery detection. $^\dagger$: in DeeperForensics-1.0~\cite{jiang2020deeperforensics10}, each fake video is randomly perturbed for augmentation, and the perturbed videos are counted as new fake videos. Ignoring perturbation, DeeperForensics-1.0 only contains {1,000} unique fake videos, much fewer than the {10,000} in \ourdataset.}
	\label{table:dataset}
	\vspace{-12pt}
\end{table}

\vspace{-2pt}
\section{Related Work}
\vspace{-1pt}
\noindent\textbf{Existing Face Forensics Datasets.}
Being the foundations of more advanced techniques, the pursuit of better datasets has attracted substantial research interest in the area of face forgery detection (see Table\!~\ref{table:dataset}). Early attempts$_{\!}$ can$_{\!}$ be traced back to$_{\!}$ MICC-F2000\!~\cite{amerini2011sift} and DSI-1\!~\cite{de2013exposing}, in which faces were manipulated in still images under strictly constrained conditions. In recent years, great efforts have been devoted to establishing video-based datasets, such as UADFV\!~\cite{yang2019exposing}, DF-TIMIT\!~\cite{korshunov2018deepfakes}, FaceForensics++\!~\cite{rossler2019faceforensics}, DFDC\!~\cite{dolhansky2020deepfake}, Celeb-DF~\cite{li2020celeb}, VideoForensicsHQ~\cite{fox2020videoforensicshq} and DeeperForensics-1.0\!~\cite{jiang2020deeperforensics10}. With constantly upgraded face forgery techniques (\eg, FaceSwap\!~\cite{faceswap}, NeuralTextures\!~\cite{thies2019deferred}, FaceShifter\!~\cite{li2019faceshifter}, NVP\!~\cite{thies2019neural}), videos of forged faces in some datasets seem deceptively real to the human eye. These datasets have undoubtedly advanced this field. Nonetheless, they are still limited in that most videos come from simple scenarios with only one or two identities. Therefore, benchmarking algorithms on these datasets is not sufficient for measuring their performance in practical scenarios.

To address the limitations of previous datasets, we introduce \ourdataset~which targets at \textit{multi-person} face forgery detection. \ourdataset~is unique in its real-world complexity (\ie, it covers multiple individuals), high-fidelity manipulation (\ie, its quality is guaranteed by a model-agnostic quality assessment network, Q-Net) and scalability (\ie, it is constructed in a unconstrained, automatic condition).



\noindent\textbf{Neural Face Synthesis and Face Forensics Dataset Construction.}
Neural face manipulation has been a long-standing research topic in computer vision and computer graphics for over two decades\!~\cite{nguyen2019deep}. The very first work, \ie, Video Rewrite\!~\cite{bregler1997video}, automatically synthesizes human faces with proper lip sync to a given audio signal. Introduced later, face swapping systems~\cite{faceswap,thies2015real,thies2016face2face} typically follow computer graphics pipelines, fitting a parametric 3D face model to target faces for manipulation. However, the performance of these methods relies heavily on the quality of the 3D model. Some alternatives~\cite{kim2018deep,thies2019deferred} thus alleviate this by combining graphics pipelines with learnable components, which can use imperfect 3D models for synthesis. More recently, GAN-based models~\cite{deepfake,wu2018reenactgan,nirkin2019fsgan,li2019faceshifter,li2020celeb,jiang2020deeperforensics10} have become popular due to their concise and flexible framework, which does not require expensive manual operation or acquisition hardware. This thus enables them to be frequently engaged in constructing face forensics datasets.

Although current deepfake generators can produce high-quality manipulations when source and target faces yield strong consistency (in terms of color, pose, illumination), they easily suffer from occlusions, glasses, profile faces or sudden motions. Therefore, during face forensics dataset construction, extensive human interventions are often included to guarantee the quality of collected data, by  manually filtering out those suboptimal tampered examples~\cite{li2020celeb,jiang2020deeperforensics10}. Thus building large-scale face forensics datasets is costly and time-consuming. We instead devise a quality assessment network to post-control the face manipulation procedure. The network is trained with domain-adversarial learning on a set of automatically collected training samples and is independent from face synthesis techniques. Thus it yields high flexibility and generalization, and allows large-scale dataset construction in a labor-efficient manner.

\noindent\textbf{Face Forgery Detection.} Recently, active research has been devoted to synthetic content detection in portrait videos~\cite{tolosana2020deepfakes,verdoliva2020media}, in order to fight against the emerging threat of face swapping techniques. Early models~\cite{bianchi2012image,li2018exposing,korshunov2018deepfakes,matern2019exploiting,yang2019exposing,agarwal2019protecting} make use of hand-crafted features (\eg, blinking patterns, temporal flickering, face warping artifacts), which are manually designed to capture visual artifacts and inconsistencies generated in the fake face synthesis process. However, due to the limited representation ability of hand-designed features, they do not fit well towards more sophisticated facial manipulation techniques. With the advance of deep neural networks, recent approaches are built upon modern network architectures (\eg, Xception~\cite{chollet2017xception}, I3D~\cite{carreira2018action}), and address image-~\cite{afchar2018mesonet,yu2019attributing,rossler2019faceforensics,wang2020fakespotter,nguyen2019use,tolosana2020deepfakes} or video-level~\cite{guera2018deepfake,amerini2019deepfake,trinh2020interpretable,jiang2020deeperforensics10,masi2020two} forgery classification. Some methods~\cite{bappy2019hybrid,nguyen2019multi,li2020face,huang2020fakelocator,du2019towards,li2019zooming} further focus on fine-grained localization of manipulated regions to provide better interpretability. Additionally, frequency domain analysis~\cite{qian2020thinking,frank2020leveraging,chen2020manipulated,durall2019unmasking,durall2020watch,masi2020two}, texture statistics~\cite{liu2020global}, audio features~\cite{chugh2020not,mittal2020emotions}, and biological signals~\cite{ciftci2020fakecatcher,hernandez2020deepfakeson,qi2020deeprhythm,fernandes2019predicting} are also becoming popular for 
recognizing fake content.

Despite their success, current deep learning methods typically conduct forgery classification on trimmed videos and are prone to failing in real-world, multi-person scenarios.  Though~\cite{li2020sharp} made an initial attempt to address this, it is still confined to constrained scenarios due to the lack of large-scale$_{\!}$ datasets. In$_{\!}$ contrast, our$_{\!}$ \ourdataset$_{\!}$ enables$_{\!}$ us$_{\!}$ to$_{\!}$ make a more in-depth exploration of this new direction.




\begin{figure}[t]
	\centering
	\includegraphics[width=\linewidth]{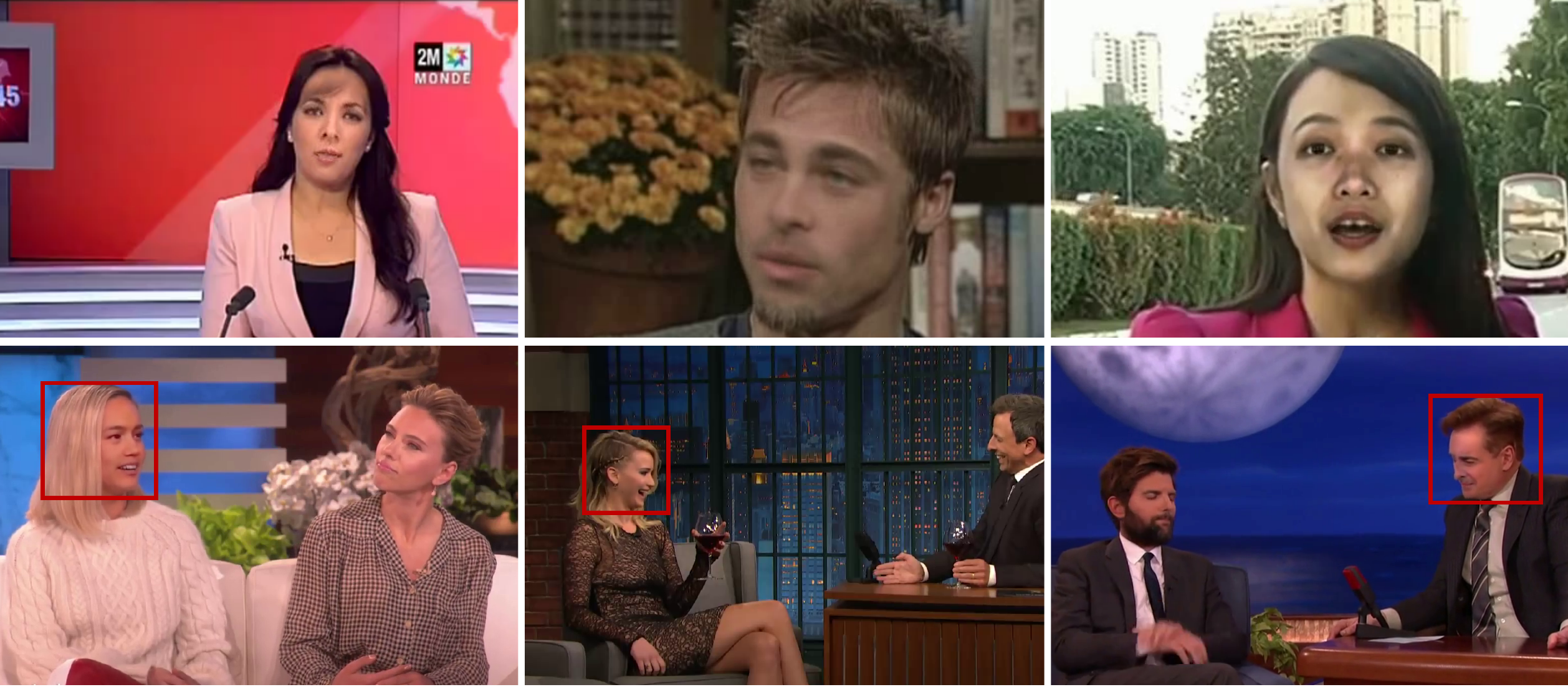}
	\vspace{-18pt}
	\captionsetup{font=small}
	\caption{\small\textbf{Exemplar frames of deepfake datasets.} From left to right: $1_{st}$ row shows tampered faces in  FaceForensics++~\cite{rossler2019faceforensics}, Celeb-DF~\cite{li2020celeb} and DeeperForensics-1.0~\cite{jiang2020deeperforensics10}; $2_{nd}$ row shows examples in \ourdataset~created by FSGAN~\cite{nirkin2019fsgan}, DeepFaceLab~\cite{petrov2020deepfacelab} and FaceSwap~\cite{faceswap}. Manipulated faces are denoted by red boxes.}
	\label{fig:example}
	\vspace{-12pt}
\end{figure}

\section{\ourdataset~Dataset}

Challenging datasets are catalysts for progress in the computer vision community. We therefore introduce \ourdataset~to provide a better benchmark and help identify conditions under which current algorithms fail, with the hope of promoting further research efforts.
Exemplars of \ourdataset~are shown in Figs.\!~\ref{fig:1} and \ref{fig:example}. In the following, we present some important aspects of \ourdataset~.

\vspace{-2pt}
\subsection{Pristine Video Collection}
\vspace{-2pt}
To align with our target of multi-person face forgery detection, we collect pristine videos in the wild, ensuring that a large number of videos contain more than one individual. We start by searching a collection of videos from \textit{YouTube} based on diverse keyword queries. To alleviate selection bias, the search is conducted by $10$ people with self-chosen queries in different languages. For video quality, we only download high-resolution videos ($480$p or higher), yielding a total of $4,\!000$ raw videos. Then, we split each video into four uniform clips, and randomly select one $12$-second sequence from each. We filter out static or crowded sequences, as well as sequences containing few human faces. This results around $12,\!000$ sequences, which we used as pristine videos for facial manipulation.


\vspace{-2pt}
\subsection{Facial Manipulation Procedure}\label{sec:manipulation}
\vspace{-2pt}
For face swapping, we randomly select two videos from the pristine collection, \ie, a \textit{target} video in which a target face will be replaced, and a \textit{source} video providing the identity of a source face that will be swapped onto the target face. Since both the source and target videos contain multiple identities, we pre-process them with off-the-shelf face detection and tracking algorithms~\cite{li2019dsfd,wojke2017simple} to obtain a set of face tracklets. We then select the tracklets with the longest duration and highest resolution for swapping. To enrich the diversity of manipulated videos, we create each video with one of three face swapping methods, including two learning-based methods (DeepFaceLab~\cite{petrov2020deepfacelab} and FSGAN~\cite{nirkin2019fsgan}), and one graphic-based method (FaceSwap~\cite{faceswap}). Though these methods can produce compelling results, they still show weaknesses under varying conditions. For example, DeepFaceLab performs poorly in the presence of glasses and extreme poses, and FSGAN is weak in maintaining an even skin tone in dark scenes. Instead of previous works~\cite{li2020celeb,jiang2020deeperforensics10} involving dense human interventions in dataset construction, we design a fully automatic procedure so that our dataset can be easily scaled. We develop a quality assessment network to quantitatively score each manipulated face, and discard the synthetic faces with low scores (see \figref{fig:q}\!~(b)), \ie, only $10,000$ fake videos with high quality scores and from different pristine videos are selected to build \ourdataset. The network allows us to build \ourdataset~with low human cost.  For conciseness, we defer the discussion of the quality control procedure to~\S\ref{sec:q}, after we have provided all the necessary details of \ourdataset~\!.


\vspace{-2pt}
\subsection{Dataset Features and Statistics}\label{sec:stat}
\vspace{-2pt}
To offer deeper insights into \ourdataset, we next discuss its various attractive properties and  descriptive  statistics.

\begin{figure}[t]
	\centering
	\vspace{-3pt}
	\includegraphics[width=\linewidth]{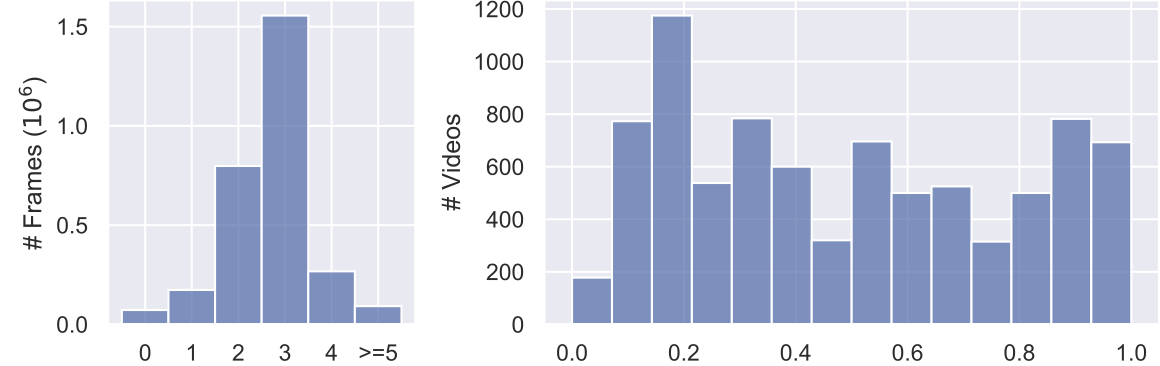}
	\put(-235,-9){\footnotesize(a) Number of faces per frame}
	\put(-127,-9){\footnotesize(b) Histogram of fake faces per video}
	\vspace{-3pt}
	\captionsetup{font=small}
	\caption{\small\textbf{$_{\!}$Statistics$_{\!}$ of$_{\!}$ \ourdataset~}(\S\ref{sec:stat}).$_{\!}$ (a)$_{\!}$ Distribution of$_{\!}$ face$_{\!}$ number per frame.\!~(b)$_{\!}$ Ratio$_{\!}$ distribution$_{\!}$ of tampered$_{\!}$ faces.}
	\label{fig:hist}
	\vspace{-14pt}
\end{figure}

\noindent\textbf{Real-World Complexity.}
Existing datasets fall in short of containing just one or two identities in each video (see Table\!~\ref{table:dataset}), which does not accurately reflect the distribution in the real world. However, \ourdataset~is designed to involve more human faces ($1$--$15$ per frame, $3.15$ on average). The distribution of face number in each frame is shown in \figref{fig:hist}\!~(a). Another challenge \ourdataset~provides is that each video contains both real and fake faces. We analyze the ratio of the number of tampered faces against the number of all faces in each video in \figref{fig:hist}\!~(b). As seen, in many videos, only a small percentage of faces are manipulated. These statistics are more representative of real world-applications and allow for in-depth benchmark analysis.

\begin{figure}[t]
	\centering
	\vspace{-3pt}
	\includegraphics[width=0.75\linewidth]{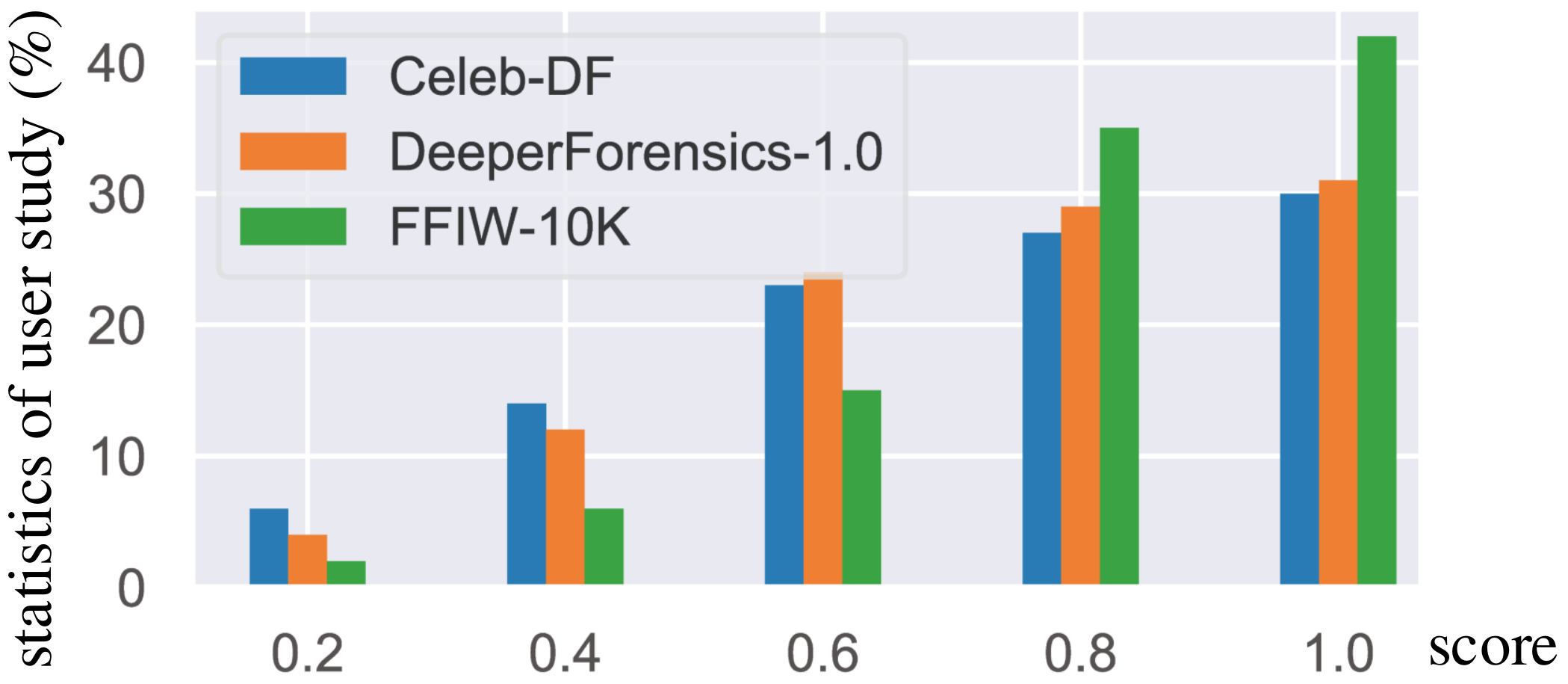}
	\vspace{-6pt}
	\captionsetup{font=small}
	\caption{\small\textbf{Statistics of user study on data fidelity}~(\S\ref{sec:stat}).}
	\label{fig:user-study}
	\vspace{-14pt}
\end{figure}

\noindent\textbf{High Fidelity with Low Human Cost.}
We organize a user study to verify the quality of \ourdataset. Specifically, a total of $50$ computer science students are invited to assess the realness of synthetic videos in \ourdataset~as well as two previous high-quality datasets (\ie, Celeb-DF~\cite{li2020celeb} and DeeperForensics-1.0~\cite{jiang2020deeperforensics10}). Following\!~\cite{jiang2020deeperforensics10}, we randomly select $30$ videos from each dataset and prepare a web-based platform to play each video once to the participants. Each participant is asked to score each video at five levels ($0.2$ -- clearly fake, $0.4$ -- fake, $0.6$ -- borderline, $0.8$ -- real, $1.0$ -- clearly real). For each video, we average the scores of all users as the final score. \figref{fig:user-study} shows the results. As seen, more videos in \ourdataset~are rated as `real' and `clearly real' than in the other two datasets. This can be attributed to: i) the intrinsic difficulties of multi-person face forgery detection; and ii) the effectiveness of the Q-Net in  quality control.

\noindent\textbf{Large Scale.}
\!As shown in~Table~\ref{table:dataset}, \ourdataset~consists of 10K synthetic as well as 10K real videos, with about $33$ hours and more than $7.2$M frames in total. Note that the number of \textit{unique} fake videos in \ourdataset~is ten orders of magnitude larger than DeeperForensics-1.0, which treats a manipulated video and its distorted versions (perturbed by Gaussian blur, JPEG compression, \etc) as different videos.

\noindent\textbf{Dataset Annotation.} For completeness, \ourdataset~provides both face-level and video-level labels; a total of $3.2$M real faces and $1.1$M fake faces of $3.6$K persons are annotated. Note that our method explores only the usage of video-level labels, addressing high utility in practical applications.

\noindent\textbf{Dataset Split.}
We split \ourdataset~into separate \texttt{train}, \texttt{val} and \texttt{test} sets. Following random selection of pristine video clips, we arrive at a unique split consisting of $16,\!000$ training, $500$ validation, and $3,\!500$ test videos. In each split, each fake video is companied with its  real video.


	\vspace{-2pt}
\section{Domain-Adversarial Quality Control}\label{sec:q}
	\vspace{-2pt}
\noindent\textbf{Domain-Adversarial Quality Assessment Network (Q-Net).} As mentioned in \S\ref{sec:manipulation}, for facilitating dataset construction, we design a Q-Net $\mathcal{F}^{Q\!}$ (VGG16-based) that automatically evaluates the quality of each swapped face and hence allows us to effortless filter out low-fidelity faces. The main challenge here is that it is hard to collect precise quality annotations directly from the face swapping algorithms (\ie, DeepFaceLab\!~\cite{petrov2020deepfacelab}, FSGAN\!~\cite{nirkin2019fsgan}, FaceSwap\!~\cite{faceswap}).
Inspired by~\cite{gu2020giqa}, we collect data in a semi-supervised way. Our algorithm
is built on the observation that, for most generative models, the quality of their synthesized images progressively improves as the training continues. This enables us to collect face images generated by various unconditional generative models\!~(\ie, StyleGAN~\cite{karras2019style}, StyleGAN2~\cite{karras2020analyzing}, PGGAN~\cite{karras2017progressive}) in different iterations and use the corresponding iteration number as the pseudo groundtruth quality score. Specifically, for each generated face $I_i$, the pseudo score is defined as: $s_i\!=\!0.9\!\times\!n/N$, where $n$ and $N$ indicate the iteration number and the maximum iteration, respectively. We train each model~\cite{karras2019style,karras2020analyzing,karras2017progressive} on FFHQ\!~\cite{karras2019style} for $N\!=\!5,\!000$ iterations and select $20$ images per iteration, leading to a total of $300,\!000$ training samples $\{I_i, s_i\}_i$.

\begin{figure}[t]
	\centering
	\includegraphics[width=\linewidth]{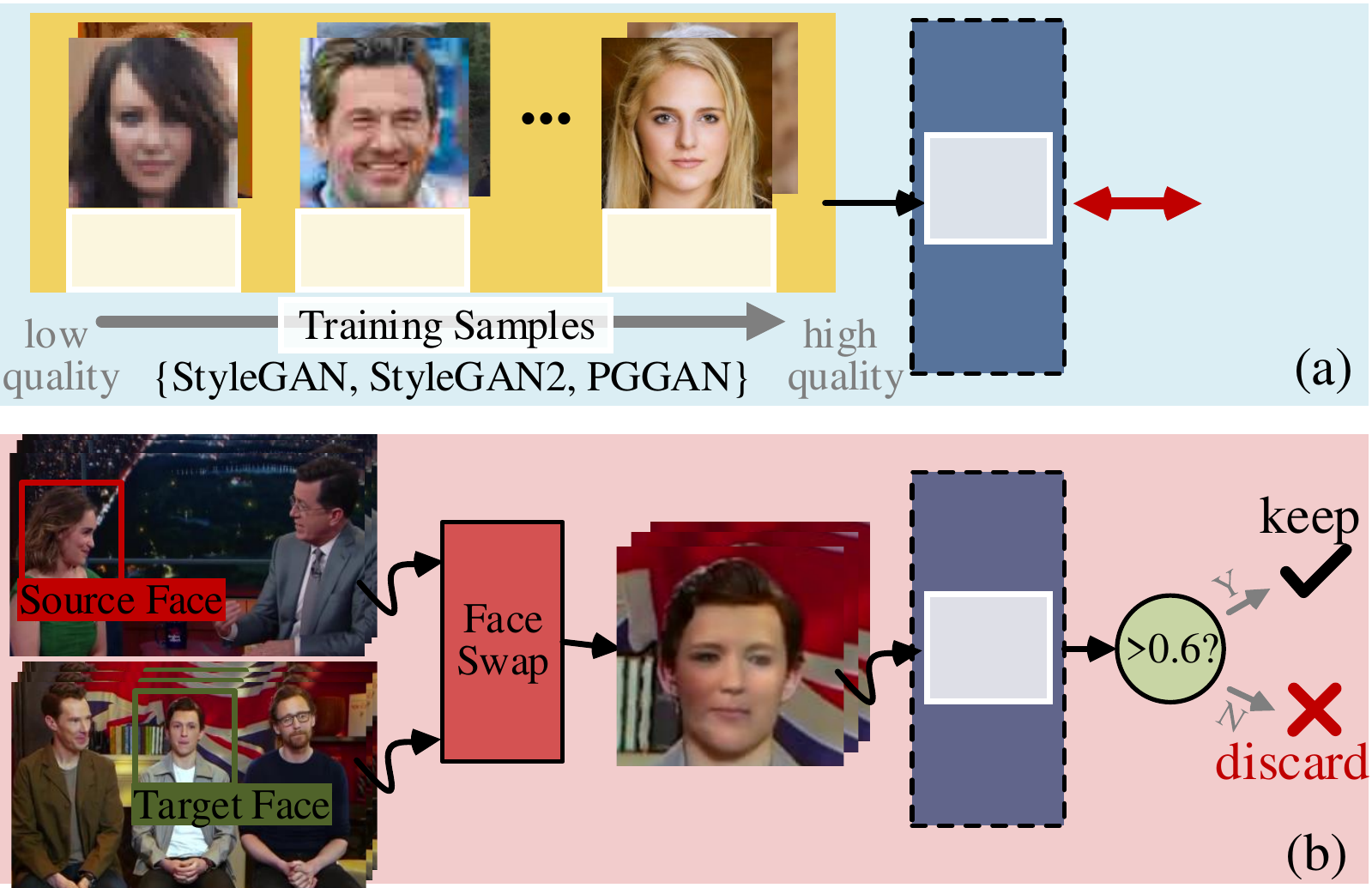}
	\put(-53,53){\footnotesize $\{\hat{s}_t\}_t$}
	\put(-100,13){\footnotesize $\{\tau_t\}_t$}
	\put(-225,110){\scriptsize $n\!=\!100$}
	\put(-225,104){\scriptsize $s\!=\!\text{0.018}$}
	\put(-184,110){\scriptsize $n\!\!=\!\!1000\!\!$}
	\put(-184,104){\scriptsize $s\!=\!0.18$}
	\put(-130,110){\scriptsize $n\!\!=\!\!5000$}
	\put(-130,104){\scriptsize $s\!=\!0.9$}
	\put(-25,115){\footnotesize $\mathcal{L}_Q$}
    \put(-32,105){\footnotesize (Eq.\!~\eqref{eq:q})}
	\put(-72,117){\footnotesize $\mathcal{F}^Q$}
	\put(-72,37){\footnotesize $\mathcal{F}^Q$}
	\vspace{-5pt}
	\captionsetup{font=small}
	\caption{\small (a) Training stage of Q-Net. (b) Facial manipulation and Q-Net based quality control during the construction of \ourdataset.}
	\label{fig:q}
	\vspace{-14pt}
\end{figure}
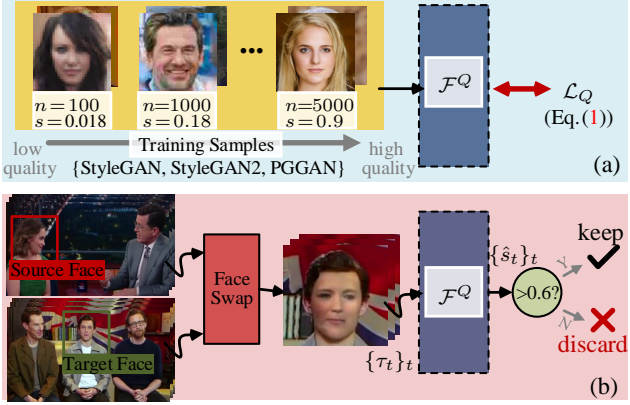

Directly learning on the collected data is insufficient, since the Q-Net may overfit to specific artifacts of different generative models~\cite{yu2019attributing,durall2020watch}, rather than learning discriminative and informative representations for realism perception. This will lead to poor generalization when tested on \ourdataset~due to the \textit{domain shift} between the training and testing distributions. To address this issue, we introduce domain-adversarial regularization~\cite{ganin2016domain} to encourage domain-invariant feature learning  in the course of the optimization. Concretely, we assign each training sample $I_i$ an extra domain label $d_i\!\in\!\{\text{StyleGAN}, \text{StyleGAN2},$ $\text{PGGAN}\}$. For each $I_i$, we let $\mathcal{F}^{Q\!}$ predict both the quality score and domain label. As shown in \figref{fig:q}\!~(a),  with all the training samples $\{I_i, s_i, d_i\}_i$, $\mathcal{F}^{Q\!}$ is learned by minimizing:
\vspace{-6pt}
\begin{equation}\small\label{eq:q}
\mathcal{L}_Q = \sum\nolimits_i \mathcal{L}_{{l_1}}(\hat{s}_i, s_i) - \alpha\mathcal{L}_{\text{CE}}(\hat{d}_i, d_i),
\vspace{-1pt}
\end{equation}
where $\mathcal{L}_{{l_1}}\!$ and $\mathcal{L}_{\text{CE}}\!$ are $l_1\!$ and cross-entropy losses, respectively, $\hat{s}$ and $\hat{d}$ indicate the estimated quality score and domain label, respectively, and $\alpha\!>\!0$.  Through the domain-adversarial regularization\!~(\ie, $- \alpha\mathcal{L}_{\text{CE}}(\hat{d}_i, d_i)$), our Q-Net is enforced to learn discriminative feature  representations that are invariant to the change of data distributions, hence gaining improved generalization ability.



To evaluate $\mathcal{F}^{Q\!}$, we carry out a user study over a set of images to measure the consistency between model predictions and human assessments (see \textit{supplementary}). The results from the user study confirm our Q-Net is effective.

\noindent\textbf{Q-Net based Quality Control.}
\!After training, Q-Net is employed to automate the construction of \ourdataset. For each swapped face $\tau$, created by DeepFaceLab, FSGAN or FaceSwap, we compute its quality score through Q-Net: $\hat{s}\!=\!\mathcal{F}^{Q\!}(\tau)$. Then, for each manipulated face tracklet, it will be preserved only if its quality score, averaged over the swapped faces it contains, is larger than $0.6$ (see \figref{fig:q}\!~(b)).

\begin{figure*}[t]
	\centering
\vspace{-3pt}
	\includegraphics[width=\linewidth]{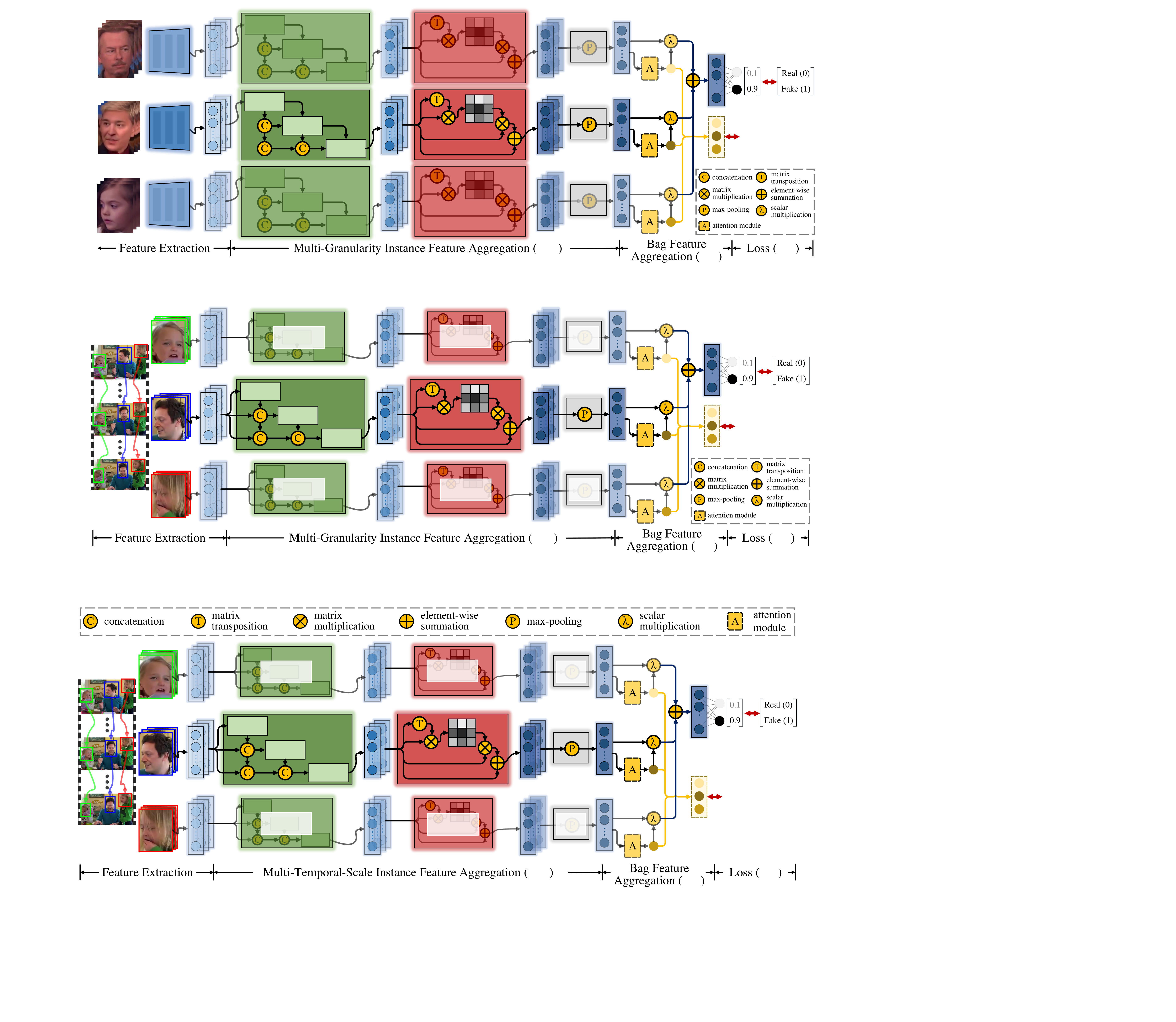}
	\put(-425,118){\scriptsize $\bm{X}\!\in\!\mathbb{R}^{D\!\times\!T}$}
	\put(-306,118){\scriptsize $\bm{S}\!\in\!\mathbb{R}^{D\!\times\!T}$}
	\put(-196,117){\scriptsize $\bm{L}\!\in\!\mathbb{R}^{D\!\times\!T}$}
	\put(-146,117){\scriptsize $\bm{Y}\!\in\!\mathbb{R}^{D}$}
	\put(-186,8.5){\scriptsize $\S\ref{sec:instance-aggregation}$}
	\put(-81,3){\scriptsize $\S\ref{sec:bag-aggregation}$}
	\put(-28,8.5){\scriptsize $\S\ref{sec:loss}$}
	\put(-391,109){\scriptsize $\mathcal{F}_1^{\text{atr\_conv}}$}
	\put(-365,93){\scriptsize $\mathcal{F}_2^{\text{atr\_conv}}$}
	\put(-335,77){\scriptsize $\mathcal{F}_3^{\text{atr\_conv}}$}
	\put(-235,90){\scriptsize $\bm{A}$}
	\put(-370,147){\scriptsize $\mathcal{F}^s$\!~(Eq.~\eqref{eq:4})}
	\put(-370,43){\scriptsize $\mathcal{F}^s$\!~(Eq.~\eqref{eq:4})}
	\put(-255,148){\scriptsize $\mathcal{F}^l$\!~(Eq.~\eqref{eq:5})}
	\put(-255,42){\scriptsize $\mathcal{F}^l$\!~(Eq.~\eqref{eq:5})}
	\put(-161,145){\scriptsize$\mathcal{F}^g$}
	\put(-161,41){\scriptsize$\mathcal{F}^g$}
	\put(-105,74){\scriptsize$a_2$}
	\put(-105,22){\scriptsize$\color{ggray}{a_3}$}
	\put(-105,126){\scriptsize$\color{ggray}{a_1}$}
	\put(-82,80){\scriptsize$\bm{a}_{\mathcal{V}\!}\!\in\![0,1]^K$}
	\put(-48,61){\scriptsize$\mathcal{L}_{\text{Sparsity}}$}
	\put(-39,105){\scriptsize$\mathcal{L}_{\text{CE}}$}
	\put(-82,143){\scriptsize$\bm{O}_{\mathcal{V}\!}\!\in\!\mathbb{R}^D$}
	\put(-47,134){\scriptsize$\hat{l}_{\mathcal{V}}$}
	\put(-19,134){\scriptsize$l_{\mathcal{V}}$}
	\put(-438,166){\scriptsize$\Gamma_1$}
	\put(-438,113){\scriptsize$\Gamma_2$}
	\put(-438,59){\scriptsize$\Gamma_3$}
	\put(-463,37){\scriptsize$\mathcal{V}$}
	\vspace{-8pt}
	\captionsetup{font=small}
	\caption{\small\textbf{Framework of the proposed discriminative attention model} (\S\ref{sec:5}) for face forgery detection in multi-person scenarios.}
	\label{fig:framework}
	\vspace{-14pt}
\end{figure*}
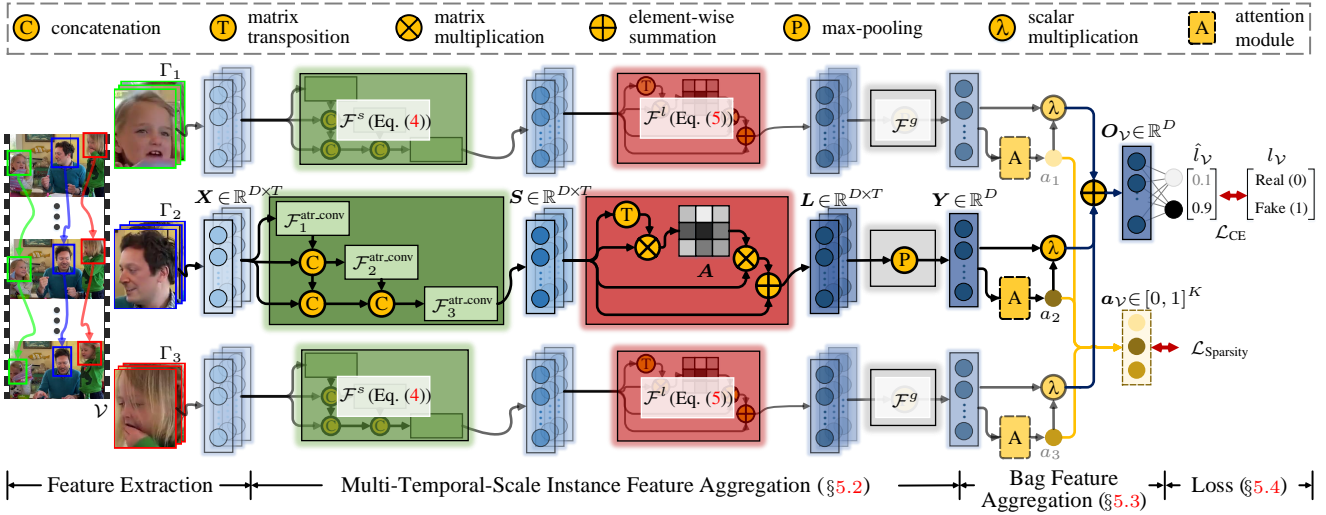

\vspace{-2pt}
\section{Face Forgery Detection Framework}\label{sec:5}
\vspace{-2pt}
\subsection{Discriminative Attention Model}
\vspace{-2pt}
In this section, we elaborate on our discriminative attention model for multi-person face forgery detection, which falls into the multiple instance learning (MIL) regime~\cite{dietterich1997solving,maron1998framework}. In MIL, labels are associated with groups of instances (or \textit{bags}), while instance labels are unobserved. The learning procedure aims to combine instance knowledge and predict labels on the bag level. In our problem, each video $\mathcal{V}$ corresponds to a bag, with its class label $l_{\mathcal{V}}\!\in\!\{\text{fake}, \text{real}\}$. A bag consists of $K$ instances with unknown labels, each of which is a tracklet of faces, obtained by~\cite{li2019dsfd,wojke2017simple}. We then formulate MIL-based face forgery detection as:
\vspace{-4pt}
\begin{equation}\small\label{eq:mil}
\begin{aligned}
\max_{\bm a_{\mathcal{V}}\in[0,1]^K}\log p({{l}}_{\mathcal{V}}|\{\bm{Y}_k\}_{k=1}^K, \bm a_{\mathcal{V}}) + \log p(\bm a_{\mathcal{V}}),
\end{aligned}
\vspace{-2pt}
\end{equation}
where $\bm{Y}_{k\!}$ denotes the representation of the $k$-th tracklet instance, and $\bm a_{\mathcal{V}}$ is a tracklet-aware attention vector, in which each value  measures the likelihood of the corresponding tracklet being fake.  The first term $\log p({{l}}_{\mathcal{V}}|\{\bm{Y}_i\}_{i=1}^K, \bm a_{\mathcal{V}}) $ prefers $\bm a_{\mathcal{V}}$ with high discriminative capacity for classification, while the second term  $\log p(\bm a_{\mathcal{V}})$ models the prior distribution of $\bm a_{\mathcal{V}}$. With Eq.\!~\eqref{eq:mil}, we design a \textit{multi-temporal-scale instance feature aggregation} module (\S\ref{sec:instance-aggregation}) to learn instance representations $\{\bm{Y}_k\}_{k=1}^K$, an \textit{attention-based bag feature aggregation} module (\S\ref{sec:bag-aggregation}) to fuse $\{\bm{Y}_k\}_{k=1}^K$ into a video descriptor according to $\bm a_{\mathcal{V}}$, and a \textit{sparse attention regularization loss} (\S\ref{sec:loss}) to model the distribution of $\bm a_{\mathcal{V}}$.





\subsection{Multi-Temporal-Scale Instance Feature Aggregation}\label{sec:instance-aggregation}
\vspace{-1pt}
Let us denote $\Gamma\!=\!\{\tau_1, \ldots, \tau_T\}$ as a tracklet instance with $T$ face regions detected from ${\mathcal{V}}$, each face region $\tau_t$ is represented by a feature vector ${\bm{x}}_t\!\in\!\mathbb{R}^{D}$.  Here we aim to learn a compact representation $\bm{Y}$ for $\Gamma$:
\vspace{-2pt}
\begin{equation}\small
{{\bm{Y}}} = \mathcal{F}({\bm{x}}_1, \ldots, {\bm{x}}_T)\in\mathbb{R}^D,
\vspace{-3pt}
\end{equation}
where the aggregation function $\mathcal{F}$ can be naturally implemented as a global pooling operation\!~(\eg, max-pooling, average pooling or log-sum-exponential pooling\!~\cite{boyd2004convex}) over all the input features. However, global statistics cannot describe rich relations among different face regions, especially the \textit{temporal order} within the tracklet, which are informative for recognizing temporal inconsistency\!~(\eg, eye blinking patterns\!~\cite{li2018ictu}, temporal artifacts\!~\cite{guera2018deepfake}) in manipulated face sequences. We thus propose a multi-temporal-scale feature aggregation module for more discriminative instance representation learning. Formally, suppose ${\bm{X}}\!=\![{\bm{x}}_1, \ldots, {\bm{x}}_T]\!\in\!\mathbb{R}^{D\times T}$ be the raw tracklet representation in matrix form. $\mathcal{F}$ is achieved by a sequence of short-term $\mathcal{F}^s$, long-term $\mathcal{F}^l$ and global $\mathcal{F}^g$ aggregation operations (see \figref{fig:framework}):
\par\nobreak
\vspace{-13pt}
{\small\setlength{\belowdisplayskip}{4pt}
\begin{align}
\textit{short-term aggregation:}& ~~~ {\bm{S}} = \mathcal{F}^s({\bm{X}}) \in \mathbb{R}^{D\times T}, \label{eq:4}\\
\textit{long-term aggregation:}& ~~~ {\bm{L}} = \mathcal{F}^l({\bm{S}}) \in \mathbb{R}^{D\times T}, \label{eq:5}\\
\textit{global aggregation:} & ~~~ {\bm{Y}} \!= \mathcal{F}^g({\bm{L}}) \in \mathbb{R}^D. \label{eq:6}
\end{align}}%
Here, $\mathcal{F}^g$ denotes max-pooling. $\bm{S}$ and ${\bm{L}}$ are intermediate features after short-term $\mathcal{F}^s$ and long-term $\mathcal{F}^l$ aggregation operations, respectively, which will be detailed later.

\noindent\textbf{Short-Term Feature Aggregation.}
We propose a densely connected dilated temporal convolution module to achieve $\mathcal{F}^s$ in Eq.\!~\eqref{eq:4}. The module combines the advantages of atrous convolution\!~\cite{chen2014semantic} and dense connectivity\!~\cite{huang2017densely} to effectively enlarge the field of view of filters to capture large temporal context. Specifically, $\mathcal{F}^s$ is a stack of $L$ atrous convolutional layers, \ie, $\{\mathcal{F}^{\text{atr\_conv}}_l\}_{l=1}^L$, where the dilation rate $r_l$ is increased layer by layer. Each $\mathcal{F}^{\text{atr\_conv}}_l$ takes the concatenated features of all proceeding layers, $[\bm{S}_0, \bm{S}_1, \ldots, \bm{S}_{l-1}]\!\in\!\mathbb{R}^{(l\times D)\times T}$ as inputs, and outputs:
\vspace{-2pt}
\begin{equation}\small\label{eq:atr}
\begin{aligned}
\bm{S}_l= \mathcal{F}^{\text{atr\_conv}}_l([\bm{S}_0, \bm{S}_1, \ldots, \bm{S}_{l-1}])\in\mathbb{R}^{D\times T},
\end{aligned}
\vspace{-2pt}
\end{equation}
where $\bm{S}_{0}\!=\!\bm{X}$.
Here,
$\mathcal{F}^{\text{atr\_conv}\!}$ is able to efficiently capture temporal patterns over a relatively wide range without drastically
increasing the number of parameters. The dense connection structure enables gradually assembling more temporal cues from different layers. Therefore,  with a large receptive field, $\mathcal{F}^s$ finally produces a powerful, short-term descriptor $\bm{S}$ for the tracklet $\Gamma$, by comprehensively modeling and fusing context over different local temporal scales.


In practice, each $\mathcal{F}^{\text{atr\_conv}}_l$ is implemented by: \!\texttt{bn}-\texttt{relu}-\texttt{conv}($1\!\times\!1$)-\texttt{bn}-\texttt{relu}-\texttt{conv}($3\!\times\!3,r_l$)-\texttt{bn}-\texttt{conv}($1\!\times\!1$). Here, the first $1\!\times\!1$ \texttt{conv} reduces the feature dimension to $(l\!\times\!D)/4$ for computational efficiency, the $3\!\times\!3$ \texttt{conv} with dilation rate $r_l$ facilitates multi-scale feature learning, and the second $1\!\times\!1$ \texttt{conv} outputs the feature $\bm{S}_l\!\in\!\mathbb{R}^{D\!\times\!T}$ at $l$-th layer. We use $L\!=\!3$ layers of dilated convolution with rates $r\!=\!\{1,2,4\}$, respectively, as shown in \figref{fig:framework}.



\noindent\textbf{Long-Term Feature Aggregation.}
\!In addition to the short-term temporal context learning, we conduct long-term context aggregation $\mathcal{F}^l$\!~(Eq.\!~\eqref{eq:5}) over short-term feature $\bm{S}$ to learn a non-local informative representation for the tracklet $\Gamma$. Specifically, we employ self-attention\!~\cite{vaswani2017attention,wang2018non} to model the long-range, multi-level dependencies among temporal features in ${\bm{S}}$\!~(see \figref{fig:framework}). We first compute the normalized correlation between each pair of temporal feature vectors in ${\bm{S}}\!=\![{\bm{s}}_1, \ldots, {\bm{s}}_T]\!\in\!\mathbb{R}^{D\times T}$ through pairwise dot product:
\vspace{-2pt}
\begin{equation}\small\label{eq:sa}
\begin{aligned}
{\bm{A}} &= \texttt{softmax}({\bm{S}}^\top{\bm{S}})\\&= \texttt{softmax}([{\bm{s}}_1, \ldots, {\bm{s}}_T]^\top[{\bm{s}}_1, \ldots, {\bm{s}}_T]) \in [0,1]^{T\times T}.
\end{aligned}
\vspace{-2pt}
\end{equation}
The affinity matrix $\bm{A}$ stores similarity scores corresponding to all pairs of features in $\bm{S}$, \ie, the $(i,j)$-th element of $\bm{A}$ gives the similarity between ${\bm{s}}_i$ and ${\bm{s}}_j$. $\texttt{softmax}(\cdot)$ normalizes each column of the input. Next, attention summaries are computed as ${\bm{S}}{\bm{A}}\!\in\!\mathbb{R}^{D\times T}$, and used to generate the long-term descriptor ${\bm{L}}$ for $\Gamma$ in a residual form:
\vspace{-4pt}
\begin{equation}\small\label{eq:yl}
\begin{aligned}
{\bm{L}}  = \mathcal{F}^l({\bm{S}}) = {\bm{S}}{\bm{A}} + {\bm{S}} \in \mathbb{R}^{D\times T}.
\end{aligned}
\vspace{-4pt}
\end{equation}
Thus $\bm{L}$ encodes both the long-term ${\bm{S}}{\bm{A}}$ and short-term information ${\bm{S}}$, with enhanced representability.


\subsection{Attention-Based Bag Feature Aggregation}\label{sec:bag-aggregation}
After applying max-pooling based global aggregation $\mathcal{F}^g$ over $\bm{L}$ (Eq.\!~\eqref{eq:6}), we get a compact and discriminative representation $\bm{Y}\!\in\!\mathbb{R}^{D}$ for each tracklet $\Gamma$. For video $\mathcal{V}$, all the $K$ detected tracklets form a bag. We further adaptively aggregate all instance features $\{{\bm{Y}}_k\}_{k=1}^K$ into a global bag-level representation, using learnable attention:
\vspace{-4pt}
\begin{equation}\small\label{eq:bag}
\begin{aligned}
{\bm{O}}_{\mathcal{V}} = \sum\nolimits_{k=1}^K a_k {\bm{Y}}_k \in \mathbb{R}^{D},
\end{aligned}
\vspace{-4pt}
\end{equation}
where $\bm{a}_{\mathcal{V}} \!=\! (a_1, \ldots, a_K)\!\in\![0,1]^K$ is a vector of scalar attention weights, and each weight $a_k$ is computed by:
\vspace{-2pt}
\begin{equation}\small
a_k = \frac{\exp\{{\bm{w}}^\top\tanh({\bm{W}}^\top{\bm{Y}}_k)\}}{\sum\nolimits_{k'=1}^K \exp\{{\bm{w}}^\top\tanh({\bm{W}}^\top{\bm{Y}}_{k'})\}}\in [0,1].
\vspace{-1pt}
\end{equation}
Here ${\bm{w}}\!\!\in\!\!\mathbb{R}^{C\!}$ and ${\bm{W}}\!\!\in\!\!\mathbb{R}^{D\times C\!}$ are learnable parameters. Through the attention-aware pooling, our method enjoys \textit{high}
\textit{flexibility} to absorb faithful knowledge from representative instances for more accurate video-level classification, and \textit{better interpretability} to locate the manipulated faces according to the attention $\bm{a}_{\mathcal{V}}$. For face forgery localization, we regard a tracklet $\Gamma_k$ as fake if $a_k\!>\!0.75$. The threshold $0.75$ is determined by grid search over \ourdataset~\texttt{val}.

\subsection{Loss Function}\label{sec:loss}
Given the video-level feature representation ${\bm{O}}_{\mathcal{V}}\!\in\!\mathbb{R}^D$, a fully-connected layer is added for forgery classification. In addition, since only a sparse subset of faces are manipulated in most videos, we introduce sparse regularization over the attention vector $\bm a_{\mathcal{V}}$ to select a few most possibly tampered faces. Thus the overall training objective is defined as:
\vspace{-4pt}
\begin{equation}\small\label{eq:loss}
\begin{aligned}
\mathcal{L} = \mathcal{L}_{\text{CE}}(\hat{{l}}_{\mathcal{V}}, l_{\mathcal{V}}) + \beta \mathcal{L}_{\text{Sparsity}}(\bm a_{\mathcal{V}}).
\end{aligned}
\vspace{-3pt}
\end{equation}
Here, $\mathcal{L}_{\text{CE}}$ indicates the binary cross-entropy loss, and $\mathcal{L}_{\text{Sparsity}}$ is a sparse regularization term that is formulated as the $l_1$ norm of $\bm a_{\mathcal{V}}$, \ie, $\mathcal{L}_{\text{Sparsity}}(\bm a_{\mathcal{V}}) \!=\! \|\bm a_{\mathcal{V}}\|_1$. The coefficient $\beta\!>\!0$ controls the trade-off between the two terms.

\vspace*{-2pt}
\subsection{Implementation Details}
\vspace*{-2pt}
\noindent\textbf{Preprocessing.}
For each video, we detect human faces in each frame~\cite{li2019dsfd} and associate them across frames to obtain a set of tracks~\cite{wojke2017simple}. To incorporate more spatial context, we extend each face region by a factor of 1.2 along the width and height, uniformly resized into $224\!\times\!224$ resolution.

\noindent\textbf{Training Details.}
We employ ResNet-50~\cite{he2016deep} as the backbone network and extract features after the average pooling layer as the representation of each face ($D\!=\!2048$). The whole network is trained end-to-end using the Adam optimizer with learning rate 1e-4 and batch size 32. During training, we apply random perturbations (\eg, horizontal flipping, color jitter) on each track for data augmentation. The coefficient $\beta$ in Eq.\!~\eqref{eq:loss} is empirically set to $0.001$.

\noindent\textbf{Reproducibility.} Our model is implemented on PyTorch and trained on four NVIDIA Tesla V100 GPUs. To provide full details of our method, our codes are released.

%

\vspace*{-2pt}
\section{Experiment}
\vspace*{-1pt}
On top of \ourdataset~\!, we examine the proposed as well as representative face forgery detection methods on two tasks:$_{\!}$ face$_{\!}$ forgery$_{\!}$ classification$_{\!}$ (\S\ref{sec:ffc})$_{\!}$ and$_{\!}$ localization$_{\!}$ (\S\ref{sec:ffl}). Then, we conduct experiments for assessing cross-dataset generalization abilities of various approaches in \S\ref{sec:gen}. Finally, in \S\ref{sec:ablation}, a set of ablation studies are performed.

\noindent\textbf{Competitors.} Most previous approaches are designed for single-person scenarios, and, in practice, suffer from training divergence on \ourdataset~(due to the influence of large amounts of real faces in manipulated videos). Therefore, we train these models on \ourdataset~\texttt{train} with face-level labels. In particular, we select four frame-based (\ie, Xception\!~\cite{rossler2019faceforensics}, MesoNet\!~\cite{afchar2018mesonet}, FWA\!~\cite{li2018exposing}, PatchForensics\!~\cite{chai2020makes}) and three video-based (\ie, C3D\!~\cite{tran2015learning}, TSN\!~\cite{wang2016temporal}, I3D\!~\cite{carreira2018action}) models for comparison. Note that these models show compelling performance on existing datasets\!~\cite{rossler2019faceforensics,li2020celeb,jiang2020deeperforensics10}. In addition, S-MIL\!~\cite{li2020sharp} is employed as another baseline which can be trained using only video-level labels. All training protocols follow the original papers unless stated otherwise.

\noindent\textbf{Evaluation Protocol.}
To fairly benchmark \ourdataset~\!, we devise a unified evaluation protocol that is applicable to all the methods. In particular, each test video is first parsed into a set of face tracklets, and each approach determines the possibility of each tracklet to be fake. This is natural for video-based methods since they work on tracklets, while for frame-level methods we use the average score of all faces in each tracklet as its score. Based on the tracklet-level predictions, we compute area under the receiver operating characteristic curve (AUC) as the metric of the classification task, as well as mean average precision (mAP) for the localization task. Following conventions~\cite{rossler2019faceforensics,li2020celeb,jiang2020deeperforensics10}, we also report video-level accuracy score (ACC) for classification.

\begin{table}[t]
	\centering
	\resizebox{0.49\textwidth}{!}{
		\setlength\tabcolsep{12pt}
		\renewcommand\arraystretch{1.0}
		\begin{tabular}{r||cc|c}
			\hline\thickhline
			\rowcolor{mygray}
			&   \multicolumn{2}{c|}{classification} & {localization}  \\ \cline{2-4}
			\rowcolor{mygray}
			\multirow{-2}{*}{Methods} & ACC (\%) & AUC (\%) & mAP (\%) \\ \hline\hline			
			\multicolumn{4}{l}{{frame-based methods}: \color{blue}{using face-level labels as supervision}} \\ \hline
			Xception~\cite{rossler2019faceforensics} & 54.1 & 56.1 & 17.9\\
			MesoNet~\cite{afchar2018mesonet}& 53.8 & 55.4& 17.7 \\
			PatchForensics~\cite{chai2020makes} & 58.9& 61.6 & 18.9 \\
			FWA~\cite{li2018exposing}& 60.2 & 63.1 & 19.2 \\ \hline\hline
			\multicolumn{4}{l}{{video-based methods}: \color{blue}{using face-level labels as supervision}} \\ \hline
			TSN~\cite{wang2016temporal}& 61.1 & 62.8 & 21.7\\
			C3D~\cite{tran2015learning}& 64.3 & 65.5 & 23.9\\
			I3D~\cite{carreira2018action}& 68.8& 69.5 & 29.7 \\ \hline\hline
			\multicolumn{4}{l}{{video-based methods}: \color{blue}{using video-level labels as supervision}} \\ \hline
			S-MIL~\cite{li2020sharp}& 59.8 & 61.2 & - \\
			\textbf{Ours}& \textbf{69.4} & \textbf{70.9} & \textbf{30.8}\\
			\hline
		\end{tabular}
	}
	\captionsetup{font=small}
	\caption{\small\textbf{Quantitative results for face forgery classification and localization} on \texttt{test} set of \ourdataset~\! (\S\ref{sec:ffc} and \S\ref{sec:ffl}).}
	\label{table:result}
	\vspace*{-13pt}
\end{table}

\vspace{-1pt}
\subsection{Face Forgery Classification}\label{sec:ffc}
\vspace{-1pt}

We first investigate the classification performance of the approaches on \ourdataset~\texttt{test}. Although this task has been well studied in single-person scenarios, we observe from Table~\ref{table:result} that previous methods produce poor classification results on \ourdataset~\!, even though they are trained with face-level labels. Our model outperforms all the compared methods by a large margin. This is encouraging given that our model only accesses to video-level labels. We note that our model significantly outperforms S-MIL\!~\cite{li2020sharp}, which is also trained using video-level labels. Additionally, we can observe that the top performance in \ourdataset~is still far from being satisfactory, thus we hope that our new dataset could encourage continuous efforts in this challenging task.
\vspace*{-1pt}
\subsection{Face Forgery Localization}\label{sec:ffl}
\begin{figure}[t]
	\centering
	\includegraphics[width=\linewidth]{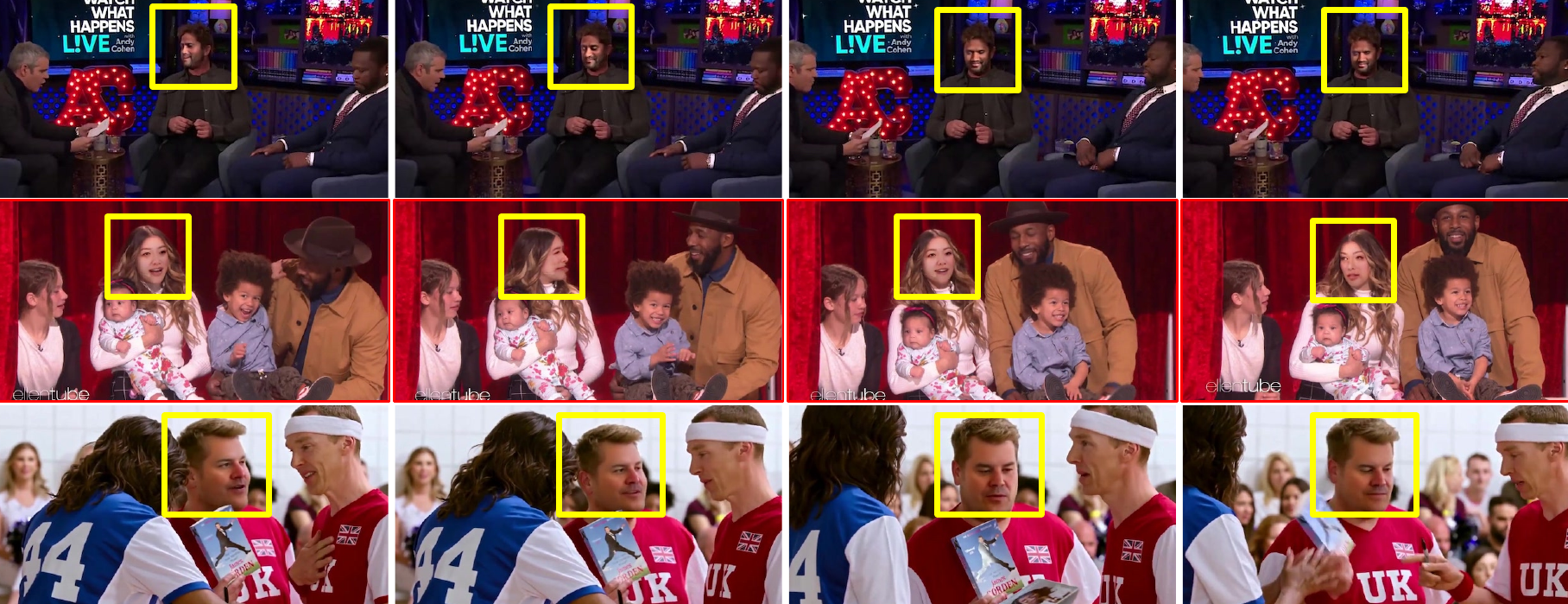}
	\vspace{-18pt}
	\captionsetup{font=small}
	\caption{\small\textbf{Visual results for face forgery localization} on \texttt{test} set of \ourdataset~(\S\ref{sec:ffl}).}
	\label{fig:result}
	\vspace{-8pt}
\end{figure}

\vspace*{-1pt}
We next analyze the performance of the approaches on face forgery localization. This task is more practical and challenging, yet is rarely explored in the literature. As shown in Table~\ref{table:result}, the video-based methods~\cite{wang2016temporal,tran2015learning,carreira2018action} consistently outperform image-based methods~\cite{rossler2019faceforensics,afchar2018mesonet,chai2020makes,li2018exposing}
in terms of mAP. Benefiting from our multi-temporal-scale feature aggregation (\S\ref{sec:instance-aggregation}) and attention-based selection (\S\ref{sec:bag-aggregation}) mechanisms, our approach achieves the best performance, even without precise, face-level supervision. Some visual results are depicted in \figref{fig:result}, showing the strong capability of our model in isolating high-fidelity tampered faces from complex, multi-person scenes.

\begin{table}[t]
	\centering
	\resizebox{0.47\textwidth}{!}{
		\setlength\tabcolsep{6pt}
		\renewcommand\arraystretch{1}
		\begin{tabular}{r||ccc}
			\hline\thickhline
			\rowcolor{mygray}
			&   FF++ & DFDC Preview &Celeb-DF\\
            \rowcolor{mygray}
\multirow{-2}{*}{Methods} &\cite{rossler2019faceforensics} &\cite{dolhansky2019deepfake} &\cite{li2020celeb}  \\
\hline\hline
			Xception-FF++~\cite{rossler2019faceforensics} & 99.2 & 49.9 & 48.2\\
			Capsule-FF++~\cite{nguyen2019use} & 96.6 & 53.3 & 57.5 \\
			Xception-c40-FF++~\cite{rossler2019faceforensics} & 95.5 & 69.7 & 65.5 \\
			F$^3$Net-FF++~\cite{qian2020thinking} & \textbf{99.9} & - & - \\
			Face X-ray-FF++~\cite{li2020face} & 99.2 & 73.5 & 74.8 \\
			TRN-FF++~\cite{masi2020two} & 99.1 & - &  76.7 \\ \hline
            Xception-c40-FFIW$_{10K}$ &95.7 &71.3 &66.9\\
			\textbf{Ours-FF++} & 99.5 & 72.8 & 75.3 \\
			\textbf{Ours-FFIW$_{10K}$} & 99.3 & \textbf{74.1} & \textbf{78.3} \\
			\hline
		\end{tabular}
	}
	\captionsetup{font=small}
	\caption{\small\textbf{Cross-dataset generalization evaluation for face forgery classification}, in terms of AUC (\%). See \S\ref{sec:gen} for details.}
	\label{table:generalization}
	\vspace*{-14pt}
\end{table}

\vspace*{-1pt}
\subsection{Cross-Dataset Evaluation and Generalization}\label{sec:gen}
\vspace*{-1pt}
Furthermore, we examine the cross-dataset performance of various approaches and the generalization ability of our \ourdataset~\!.$_{\!}$ As$_{\!}$ shown$_{\!}$ in$_{\!}$ Table\!~\ref{table:generalization}, all$_{\!}$ comparative$_{\!}$ methods$_{\!}$ are$_{\!}$ trained$_{\!}$ on$_{\!}$ FF++ \texttt{train}\!~\cite{rossler2019faceforensics} and evaluated on test sets of FF++\!~\cite{rossler2019faceforensics}, DFDC Preview\!~\cite{dolhansky2019deepfake}, and Celeb-DF\!~\cite{li2020celeb}, respectively. ``Ours-FF++'', also trained on FF++, produces comparable performance against other competitors, verifying the efficacy of our model. In addition, we evaluate the generalization ability of our \ourdataset~\!. We train our model as well as Xception-c40~\cite{rossler2019faceforensics} on \ourdataset~\texttt{train} and report their performance on other datasets. We see that both ``Ours-\ourdataset'' and ``Xception-c40-\ourdataset'' outperform their alternatives, \ie, ``Ours-FF++'' and ``Xception-c40-FF++'', which are trained on FF++ \texttt{train}. Hence, ``Ours-\ourdataset'' shows superior performance on DFDC Preview and Celeb-DF. These experiment results reveal that \ourdataset~has a low data bias and is well qualified to be used for training and evaluating face forgery detection models.

\vspace*{-2pt}
\subsection{Model Ablations}\label{sec:ablation}
\vspace*{-2pt}
We next conduct ablative studies of our model on \ourdataset~\texttt{test}. The results are summarized in Table~\ref{table:ablation}.


\noindent\textbf{Instance Feature Aggregation.} To study the impact of our multi-temporal-scale context aggregation (Eqs.\!~(\ref{eq:4}-\ref{eq:6})), we first develop two baselines by directly applying max- or avg-pooling over raw tracklet feature ${\bm{X}}$ to obtain a global compact descriptor ${\bm{Y}}$, without multi-temporal-scale feature learning. We can easily find that both models perform significantly worse than our full model across all metrics. This confirms that the global statistics are not eligible for encoding high-order relationships in ${\bm{X}}$, resulting in poor performance. We further separately analyze the short-term (Eq.\!~\eqref{eq:4}) and long-term (Eq.\!~\eqref{eq:5}) aggregation modules. As seen, by dropping the short-term module, the model encounters a performance drop ($70.9\%\!\rightarrow\!69.7\%$ over AUC, and $30.8\%\!\rightarrow\!29.6\%$ over mAP). A similar trend is also observed after discarding the long-term aggregation module.

\begin{table}[t]
	\centering
	\resizebox{0.49\textwidth}{!}{
		\setlength\tabcolsep{1pt}
		\renewcommand\arraystretch{1}
		\begin{tabular}{r|c||cc|c}
			\hline\thickhline
			\rowcolor{mygray}
			&   & \multicolumn{2}{c|}{{classification}} & {localization}  \\ \cline{3-4}
			\rowcolor{mygray}
			\multirow{-2}{*}{Aspect}  & 	\multirow{-2}{*}{Variants}  & ACC\!~(\%) & AUC\!~(\%) & mAP\!~(\%) \\ \hline\hline			
			\textbf{Full Model} & - & \textbf{69.4} & \textbf{70.9} & \textbf{30.8} \\ \hline
			\multirow{4}{*}{\tabincell{c}{{Instance Feature}\\{Aggregation\!~(\S\ref{sec:instance-aggregation})}} }	& max-pooling & 64.5  & 66.2  & 24.6 \\	
			& avg-pooling & 63.9 & 65.7 & 24.1 \\ \cline{2-5}
			& \textit{w/o} short-term (Eq.\!~\eqref{eq:4}) & 68.3 & 69.7 & 29.6 \\
			& \textit{w/o} long-term (Eq.\!~\eqref{eq:5}) & 69.0 & 70.4 & 30.2 \\ \hline
			
			\multirow{2}{*}{\tabincell{c}{{Bag Feature}\\{Aggregation\!~(\S\ref{sec:bag-aggregation})}} }	& max-pooling & 67.3 & 69.5 & - \\
			& avg-pooling & 64.7 & 66.8 & - \\ \hline
			Loss Function\!~(\S\ref{sec:loss}) & \textit{w/o} $\mathcal{L}_{\text{sparsity}}$ (Eq.\!~\eqref{eq:loss})  & 68.6 & 70.3 & 28.5 \\
			\hline
		\end{tabular}
	}
	\captionsetup{font=small}
	\caption{\small\textbf{Ablation study} on \texttt{test} set of \ourdataset~(see \S\ref{sec:ablation}).}
	\label{table:ablation}
	\vspace*{-16pt}
\end{table}

\noindent\textbf{Bag Feature Aggregation.} To investigate the efficacy of the learnable attention mechanism for bag feature aggregation, we compare it with two baseline models which carry out the video-level feature summarization in Eq.\!~\eqref{eq:bag} by max- and avg-pooling, respectively. We see that our attention-based aggregation mechanism brings favorable performance improvements over the baselines for classification, which can be attributed to its ability to automatically highlight the most possible instances for discrimination.

\noindent\textbf{Efficacy of $\mathcal{L}_{\text{Sparsity}}$.} At last, we study the necessity of the sparsity constraint $\mathcal{L}_{\text{Sparsity}}$ in Eq.\!~\eqref{eq:loss}. Since the term provides an appropriate modeling of the data distribution (\ie, tampered faces are sparse in manipulated videos), it contributes to great performance improvements, especially in the localization task ($28.5\%\!\rightarrow\!30.8\%$ in terms of mAP).

\vspace*{-5pt}
\section{Limitation and Discussion}
\vspace*{-3pt}
For our dataset, its difficulty is limited to the adopted face swapping algorithms. This limitation is also shared by existing datasets. Considering the rapid advance of face swapping and forgery detection techniques, it is hard to maintain a long life-span for face forensics datasets. Our domain-adversarial quality control strategy may provide a feasible solution -- one can automatically update the dataset by using more advanced deepfake techniques. For our model, it faces difficulties in the scenes with slow illumination change and stable motions. In such cases, the manipulated faces within a same tracklet usually show strong consistency and less artifacts. Thus the long-term features are less informative, easily leading to inferior performance. Hence, current study for ``forgery'' is mainly around face swapping. However, given the broader concerns about how imagery is being altered in order to influence political sphere, ``forgery'' should be explored in a larger extent, such as manipulating body movements, changing facial expressions, synthesizing realistic talking head videos, or swapping faces under controllable camera characteristics.



{\small
\bibliographystyle{ieee_fullname}
\bibliography{egbib}

\begin{thebibliography}{10}\itemsep=-1pt

\bibitem{deepfake}
Deepfakes.
\newblock \url{https://github.com/deepfakes/faceswap}.
\newblock accessed November 10, 2020.

\bibitem{faceswap}
Faceswap.
\newblock \url{https://github.com/MarekKowalski/FaceSwap/}.
\newblock accessed November 10, 2020.

\bibitem{afchar2018mesonet}
Darius Afchar, Vincent Nozick, Junichi Yamagishi, and Isao Echizen.
\newblock Mesonet: a compact facial video forgery detection network.
\newblock In {\em WIFS}, 2018.

\bibitem{agarwal2019protecting}
Shruti Agarwal, Hany Farid, Yuming Gu, Mingming He, Koki Nagano, and Hao Li.
\newblock Protecting world leaders against deep fakes.
\newblock In {\em CVPR Workshop}, 2019.

\bibitem{amerini2011sift}
Irene Amerini, Lamberto Ballan, Roberto Caldelli, Alberto Del~Bimbo, and
  Giuseppe Serra.
\newblock A sift-based forensic method for copy--move attack detection and
  transformation recovery.
\newblock {\em IEEE TIFS}, 6(3):1099--1110, 2011.

\bibitem{amerini2019deepfake}
Irene Amerini, Leonardo Galteri, Roberto Caldelli, and Alberto Del~Bimbo.
\newblock Deepfake video detection through optical flow based cnn.
\newblock In {\em ICCV Workshop}, 2019.

\bibitem{bappy2019hybrid}
Jawadul~H Bappy, Cody Simons, Lakshmanan Nataraj, BS Manjunath, and Amit~K
  Roy-Chowdhury.
\newblock Hybrid lstm and encoder--decoder architecture for detection of image
  forgeries.
\newblock {\em IEEE TIP}, 28(7):3286--3300, 2019.

\bibitem{bianchi2012image}
Tiziano Bianchi and Alessandro Piva.
\newblock Image forgery localization via block-grained analysis of jpeg
  artifacts.
\newblock {\em IEEE TIFS}, 7(3):1003--1017, 2012.

\bibitem{Google2019DeepFakeDetection}
Google~AI blog.
\newblock Contributing data to deepfake detection research.
\newblock 2019.

\bibitem{boyd2004convex}
Stephen Boyd, Stephen~P Boyd, and Lieven Vandenberghe.
\newblock {\em Convex optimization}.
\newblock Cambridge university press, 2004.

\bibitem{bregler1997video}
Christoph Bregler, Michele Covell, and Malcolm Slaney.
\newblock Video rewrite: Driving visual speech with audio.
\newblock In {\em SIGGRAPH}, 1997.

\bibitem{carreira2018action}
Jo{\~a}o Carreira, Andrew Zisserman, and Quo Vadis.
\newblock Action recognition? a new model and the kinetics dataset.
\newblock In {\em CVPR}, 2018.

\bibitem{chai2020makes}
Lucy Chai, David Bau, Ser-Nam Lim, and Phillip Isola.
\newblock What makes fake images detectable? understanding properties that
  generalize.
\newblock In {\em ECCV}, 2020.

\bibitem{chen2014semantic}
Liang-Chieh Chen, George Papandreou, Iasonas Kokkinos, Kevin Murphy, and Alan~L
  Yuille.
\newblock Semantic image segmentation with deep convolutional nets and fully
  connected crfs.
\newblock In {\em ICLR}, 2015.

\bibitem{chen2020manipulated}
Zehao Chen and Hua Yang.
\newblock Manipulated face detector: Joint spatial and frequency domain
  attention network.
\newblock {\em arXiv preprint arXiv:2005.02958}, 2020.

\bibitem{chollet2017xception}
Fran{\c{c}}ois Chollet.
\newblock Xception: Deep learning with depthwise separable convolutions.
\newblock In {\em CVPR}, 2017.

\bibitem{chugh2020not}
Komal Chugh, Parul Gupta, Abhinav Dhall, and Ramanathan Subramanian.
\newblock Not made for each other-audio-visual dissonance-based deepfake
  detection and localization.
\newblock {\em arXiv preprint arXiv:2005.14405}, 2020.

\bibitem{ciftci2020fakecatcher}
Umur~Aybars Ciftci, Ilke Demir, and Lijun Yin.
\newblock Fakecatcher: Detection of synthetic portrait videos using biological
  signals.
\newblock {\em IEEE TPAMI}, 2020.

\bibitem{de2013exposing}
Tiago~Jos{\'e} De~Carvalho, Christian Riess, Elli Angelopoulou, Helio Pedrini,
  and Anderson de Rezende~Rocha.
\newblock Exposing digital image forgeries by illumination color
  classification.
\newblock {\em IEEE TIFS}, 8(7):1182--1194, 2013.

\bibitem{dietterich1997solving}
Thomas~G Dietterich, Richard~H Lathrop, and Tom{\'a}s Lozano-P{\'e}rez.
\newblock Solving the multiple instance problem with axis-parallel rectangles.
\newblock {\em Artificial intelligence}, 89(1-2):31--71, 1997.

\bibitem{dolhansky2020deepfake}
Brian Dolhansky, Joanna Bitton, Ben Pflaum, Jikuo Lu, Russ Howes, Menglin Wang,
  and Cristian~Canton Ferrer.
\newblock The deepfake detection challenge dataset.
\newblock {\em arXiv preprint arXiv:2006.07397}, 2020.

\bibitem{dolhansky2019deepfake}
Brian Dolhansky, Russ Howes, Ben Pflaum, Nicole Baram, and Cristian~Canton
  Ferrer.
\newblock The deepfake detection challenge (dfdc) preview dataset.
\newblock {\em arXiv preprint arXiv:1910.08854}, 2019.

\bibitem{du2019towards}
Mengnan Du, Shiva Pentyala, Yuening Li, and Xia Hu.
\newblock Towards generalizable forgery detection with locality-aware
  autoencoder.
\newblock {\em arXiv preprint arXiv:1909.05999}, 2019.

\bibitem{durall2020watch}
Ricard Durall, Margret Keuper, and Janis Keuper.
\newblock Watch your up-convolution: Cnn based generative deep neural networks
  are failing to reproduce spectral distributions.
\newblock In {\em CVPR}, 2020.

\bibitem{durall2019unmasking}
Ricard Durall, Margret Keuper, Franz-Josef Pfreundt, and Janis Keuper.
\newblock Unmasking deepfakes with simple features.
\newblock {\em arXiv preprint arXiv:1911.00686}, 2019.

\bibitem{fernandes2019predicting}
Steven Fernandes, Sunny Raj, Eddy Ortiz, Iustina Vintila, Margaret Salter,
  Gordana Urosevic, and Sumit Jha.
\newblock Predicting heart rate variations of deepfake videos using neural ode.
\newblock In {\em ICCV Workshop}, 2019.

\bibitem{fox2020videoforensicshq}
Gereon Fox, Wentao Liu, Hyeongwoo Kim, Hans-Peter Seidel, Mohamed Elgharib, and
  Christian Theobalt.
\newblock Videoforensicshq: Detecting high-quality manipulated face videos.
\newblock {\em arXiv preprint arXiv:2005.10360}, 2020.

\bibitem{frank2020leveraging}
Joel Frank, Thorsten Eisenhofer, Lea Sch{\"o}nherr, Asja Fischer, Dorothea
  Kolossa, and Thorsten Holz.
\newblock Leveraging frequency analysis for deep fake image recognition.
\newblock {\em arXiv preprint arXiv:2003.08685}, 2020.

\bibitem{ganin2016domain}
Yaroslav Ganin, Evgeniya Ustinova, Hana Ajakan, Pascal Germain, Hugo
  Larochelle, Fran{\c{c}}ois Laviolette, Mario Marchand, and Victor Lempitsky.
\newblock Domain-adversarial training of neural networks.
\newblock {\em JMLR}, 17(1):2096--2030, 2016.

\bibitem{gu2020giqa}
Shuyang Gu, Jianmin Bao, Dong Chen, and Fang Wen.
\newblock Giqa: Generated image quality assessment.
\newblock In {\em ECCV}, 2020.

\bibitem{guera2018deepfake}
David G{\"u}era and Edward~J Delp.
\newblock Deepfake video detection using recurrent neural networks.
\newblock In {\em AVSS}, 2018.

\bibitem{he2016deep}
Kaiming He, Xiangyu Zhang, Shaoqing Ren, and Jian Sun.
\newblock Deep residual learning for image recognition.
\newblock In {\em CVPR}, 2016.

\bibitem{hernandez2020deepfakeson}
Javier Hernandez-Ortega, Ruben Tolosana, Julian Fierrez, and Aythami Morales.
\newblock Deepfakeson-phys: Deepfakes detection based on heart rate estimation.
\newblock {\em arXiv preprint arXiv:2010.00400}, 2020.

\bibitem{huang2017densely}
Gao Huang, Zhuang Liu, Laurens Van Der~Maaten, and Kilian~Q Weinberger.
\newblock Densely connected convolutional networks.
\newblock In {\em CVPR}, 2017.

\bibitem{huang2020fakelocator}
Yihao Huang, Felix Juefei-Xu, Run Wang, Xiaofei Xie, Lei Ma, Jianwen Li, Weikai
  Miao, Yang Liu, and Geguang Pu.
\newblock Fakelocator: Robust localization of gan-based face manipulations via
  semantic segmentation networks with bells and whistles.
\newblock {\em arXiv preprint arXiv:2001.09598}, 2020.

\bibitem{huh2018fighting}
Minyoung Huh, Andrew Liu, Andrew Owens, and Alexei~A Efros.
\newblock Fighting fake news: Image splice detection via learned
  self-consistency.
\newblock In {\em ECCV}, 2018.

\bibitem{jiang2020deeperforensics10}
Liming Jiang, Ren Li, Wayne Wu, Chen Qian, and Chen~Change Loy.
\newblock Deeperforensics-1.0: A large-scale dataset for real-world face
  forgery detection.
\newblock In {\em CVPR}, 2020.

\bibitem{karras2017progressive}
Tero Karras, Timo Aila, Samuli Laine, and Jaakko Lehtinen.
\newblock Progressive growing of gans for improved quality, stability, and
  variation.
\newblock In {\em ICLR}, 2017.

\bibitem{karras2019style}
Tero Karras, Samuli Laine, and Timo Aila.
\newblock A style-based generator architecture for generative adversarial
  networks.
\newblock In {\em CVPR}, 2019.

\bibitem{karras2020analyzing}
Tero Karras, Samuli Laine, Miika Aittala, Janne Hellsten, Jaakko Lehtinen, and
  Timo Aila.
\newblock Analyzing and improving the image quality of stylegan.
\newblock In {\em CVPR}, 2020.

\bibitem{kim2018deep}
Hyeongwoo Kim, Pablo Garrido, Ayush Tewari, Weipeng Xu, Justus Thies, Matthias
  Niessner, Patrick P{\'e}rez, Christian Richardt, Michael Zollh{\"o}fer, and
  Christian Theobalt.
\newblock Deep video portraits.
\newblock {\em ACM TOG}, 37(4):1--14, 2018.

\bibitem{korshunov2018deepfakes}
Pavel Korshunov and S{\'e}bastien Marcel.
\newblock Deepfakes: a new threat to face recognition? assessment and
  detection.
\newblock {\em arXiv preprint arXiv:1812.08685}, 2018.

\bibitem{li2019zooming}
Jia Li, Tong Shen, Wei Zhang, Hui Ren, Dan Zeng, and Tao Mei.
\newblock Zooming into face forensics: A pixel-level analysis.
\newblock {\em arXiv preprint arXiv:1912.05790}, 2019.

\bibitem{li2019dsfd}
Jian Li, Yabiao Wang, Changan Wang, Ying Tai, Jianjun Qian, Jian Yang, Chengjie
  Wang, Jilin Li, and Feiyue Huang.
\newblock Dsfd: dual shot face detector.
\newblock In {\em CVPR}, 2019.

\bibitem{li2019faceshifter}
Lingzhi Li, Jianmin Bao, Hao Yang, Dong Chen, and Fang Wen.
\newblock Advancing high fidelity identity swapping for forgery detection.
\newblock In {\em CVPR}, 2020.

\bibitem{li2020face}
Lingzhi Li, Jianmin Bao, Ting Zhang, Hao Yang, Dong Chen, Fang Wen, and Baining
  Guo.
\newblock Face x-ray for more general face forgery detection.
\newblock In {\em CVPR}, 2020.

\bibitem{li2020sharp}
Xiaodan Li, Yining Lang, Yuefeng Chen, Xiaofeng Mao, Yuan He, Shuhui Wang, Hui
  Xue, and Quan Lu.
\newblock Sharp multiple instance learning for deepfake video detection.
\newblock In {\em ACM MM}, 2020.

\bibitem{li2018ictu}
Yuezun Li, Ming-Ching Chang, and Siwei Lyu.
\newblock In ictu oculi: Exposing ai generated fake face videos by detecting
  eye blinking.
\newblock {\em arXiv preprint arXiv:1806.02877}, 2018.

\bibitem{li2018exposing}
Yuezun Li and Siwei Lyu.
\newblock Exposing deepfake videos by detecting face warping artifacts.
\newblock {\em arXiv preprint arXiv:1811.00656}, 2018.

\bibitem{li2020celeb}
Yuezun Li, Pu Sun, Honggang Qi, and Siwei Lyu.
\newblock {Celeb-DF: A Large-scale Challenging Dataset for DeepFake Forensics}.
\newblock In {\em CVPR}, 2020.

\bibitem{liu2020global}
Zhengzhe Liu, Xiaojuan Qi, and Philip~HS Torr.
\newblock Global texture enhancement for fake face detection in the wild.
\newblock In {\em CVPR}, 2020.

\bibitem{maron1998framework}
Oded Maron and Tom{\'a}s Lozano-P{\'e}rez.
\newblock A framework for multiple-instance learning.
\newblock In {\em NIPS}, 1998.

\bibitem{masi2020two}
Iacopo Masi, Aditya Killekar, Royston~Marian Mascarenhas, Shenoy~Pratik
  Gurudatt, and Wael AbdAlmageed.
\newblock Two-branch recurrent network for isolating deepfakes in videos.
\newblock In {\em ECCV}, 2020.

\bibitem{matern2019exploiting}
Falko Matern, Christian Riess, and Marc Stamminger.
\newblock Exploiting visual artifacts to expose deepfakes and face
  manipulations.
\newblock In {\em WACV Workshop}, 2019.

\bibitem{mittal2020emotions}
Trisha Mittal, Uttaran Bhattacharya, Rohan Chandra, Aniket Bera, and Dinesh
  Manocha.
\newblock Emotions don't lie: A deepfake detection method using audio-visual
  affective cues.
\newblock {\em arXiv preprint arXiv:2003.06711}, 2020.

\bibitem{nguyen2019multi}
Huy~H Nguyen, Fuming Fang, Junichi Yamagishi, and Isao Echizen.
\newblock Multi-task learning for detecting and segmenting manipulated facial
  images and videos.
\newblock {\em arXiv preprint arXiv:1906.06876}, 2019.

\bibitem{nguyen2019use}
Huy~H Nguyen, Junichi Yamagishi, and Isao Echizen.
\newblock Use of a capsule network to detect fake images and videos.
\newblock {\em arXiv preprint arXiv:1910.12467}, 2019.

\bibitem{nguyen2019deep}
Thanh~Thi Nguyen, Cuong~M Nguyen, Dung~Tien Nguyen, Duc~Thanh Nguyen, and Saeid
  Nahavandi.
\newblock Deep learning for deepfakes creation and detection.
\newblock {\em arXiv preprint arXiv:1909.11573}, 1, 2019.

\bibitem{nirkin2019fsgan}
Yuval Nirkin, Yosi Keller, and Tal Hassner.
\newblock Fsgan: Subject agnostic face swapping and reenactment.
\newblock In {\em ICCV}, 2019.

\bibitem{petrov2020deepfacelab}
Ivan Petrov, Daiheng Gao, Nikolay Chervoniy, Kunlin Liu, Sugasa Marangonda,
  Chris Um{\'e}, Jian Jiang, Luis RP, Sheng Zhang, Pingyu Wu, et~al.
\newblock Deepfacelab: A simple, flexible and extensible face swapping
  framework.
\newblock {\em arXiv preprint arXiv:2005.05535}, 2020.

\bibitem{qi2020deeprhythm}
Hua Qi, Qing Guo, Felix Juefei-Xu, Xiaofei Xie, Lei Ma, Wei Feng, Yang Liu, and
  Jianjun Zhao.
\newblock Deeprhythm: Exposing deepfakes with attentional visual heartbeat
  rhythms.
\newblock {\em arXiv preprint arXiv:2006.07634}, 2020.

\bibitem{qian2020thinking}
Yuyang Qian, Guojun Yin, Lu Sheng, Zixuan Chen, and Jing Shao.
\newblock Thinking in frequency: Face forgery detection by mining
  frequency-aware clues.
\newblock In {\em ECCV}, 2020.

\bibitem{rossler2019faceforensics}
Andreas Rossler, Davide Cozzolino, Luisa Verdoliva, Christian Riess, Justus
  Thies, and Matthias Nie{\ss}ner.
\newblock Faceforensics++: Learning to detect manipulated facial images.
\newblock In {\em ICCV}, 2019.

\bibitem{thies2019neural}
Justus Thies, Mohamed Elgharib, Ayush Tewari, Christian Theobalt, and Matthias
  Nie{\ss}ner.
\newblock Neural voice puppetry: Audio-driven facial reenactment.
\newblock In {\em ECCV}, 2019.

\bibitem{thies2019deferred}
Justus Thies, Michael Zollh{\"o}fer, and Matthias Nie{\ss}ner.
\newblock Deferred neural rendering: Image synthesis using neural textures.
\newblock {\em ACM TOG}, 38(4):1--12, 2019.

\bibitem{thies2015real}
Justus Thies, Michael Zollh{\"o}fer, Matthias Nie{\ss}ner, Levi Valgaerts, Marc
  Stamminger, and Christian Theobalt.
\newblock Real-time expression transfer for facial reenactment.
\newblock {\em ACM TOG}, 34(6):183--1, 2015.

\bibitem{thies2016face2face}
Justus Thies, Michael Zollhofer, Marc Stamminger, Christian Theobalt, and
  Matthias Nie{\ss}ner.
\newblock Face2face: Real-time face capture and reenactment of rgb videos.
\newblock In {\em CVPR}, 2016.

\bibitem{tolosana2020deepfakes}
Ruben Tolosana, Ruben Vera-Rodriguez, Julian Fierrez, Aythami Morales, and
  Javier Ortega-Garcia.
\newblock Deepfakes and beyond: A survey of face manipulation and fake
  detection.
\newblock {\em arXiv preprint arXiv:2001.00179}, 2020.

\bibitem{torralba2011unbiased}
Antonio Torralba and Alexei~A Efros.
\newblock Unbiased look at dataset bias.
\newblock In {\em CVPR}, 2011.

\bibitem{tran2015learning}
Du Tran, Lubomir Bourdev, Rob Fergus, Lorenzo Torresani, and Manohar Paluri.
\newblock Learning spatiotemporal features with 3d convolutional networks.
\newblock In {\em ICCV}, 2015.

\bibitem{trinh2020interpretable}
Loc Trinh, Michael Tsang, Sirisha Rambhatla, and Yan Liu.
\newblock Interpretable deepfake detection via dynamic prototypes.
\newblock {\em arXiv preprint arXiv:2006.15473}, 2020.

\bibitem{vaswani2017attention}
Ashish Vaswani, Noam Shazeer, Niki Parmar, Jakob Uszkoreit, Llion Jones,
  Aidan~N Gomez, {\L}ukasz Kaiser, and Illia Polosukhin.
\newblock Attention is all you need.
\newblock In {\em NIPS}, 2017.

\bibitem{verdoliva2020media}
Luisa Verdoliva.
\newblock Media forensics and deepfakes: an overview.
\newblock {\em arXiv preprint arXiv:2001.06564}, 2020.

\bibitem{wang2016temporal}
Limin Wang, Yuanjun Xiong, Zhe Wang, Yu Qiao, Dahua Lin, Xiaoou Tang, and Luc
  Van~Gool.
\newblock Temporal segment networks: Towards good practices for deep action
  recognition.
\newblock In {\em ECCV}, 2016.

\bibitem{wang2020fakespotter}
Run Wang, Felix Juefei-Xu, Lei Ma, Xiaofei Xie, Yihao Huang, Jian Wang, and
  Yang Liu.
\newblock Fakespotter: A simple yet robust baseline for spotting ai-synthesized
  fake faces.
\newblock In {\em IJCAI}, 2020.

\bibitem{wang2018non}
Xiaolong Wang, Ross Girshick, Abhinav Gupta, and Kaiming He.
\newblock Non-local neural networks.
\newblock In {\em CVPR}, 2018.

\bibitem{wojke2017simple}
Nicolai Wojke, Alex Bewley, and Dietrich Paulus.
\newblock Simple online and realtime tracking with a deep association metric.
\newblock In {\em ICIP}, 2017.

\bibitem{wu2018reenactgan}
Wayne Wu, Yunxuan Zhang, Cheng Li, Chen Qian, and Chen Change~Loy.
\newblock Reenactgan: Learning to reenact faces via boundary transfer.
\newblock In {\em ECCV}, 2018.

\bibitem{yang2019exposing}
Xin Yang, Yuezun Li, and Siwei Lyu.
\newblock Exposing deep fakes using inconsistent head poses.
\newblock In {\em ICASSP}, 2019.

\bibitem{yu2019attributing}
Ning Yu, Larry Davis, and Mario Fritz.
\newblock Attributing fake images to gans: Learning and analyzing gan
  fingerprints.
\newblock In {\em ICCV}, 2019.

\end{thebibliography}
}

\end{document}